\documentclass[3p,onecolumn,authoryear,review]{elsarticle}

\usepackage[utf8]{inputenc}
\usepackage{graphicx}
\usepackage{amsmath} 
\usepackage{amsthm}  
\usepackage{amsfonts}
\usepackage{dsfont}
\usepackage[dvipsnames]{xcolor}
\usepackage{tikz}
\usepackage[nolist,nohyperlinks]{acronym}
\usepackage[capitalize]{cleveref}\Crefname{figure}{Figure}{Figures}\Crefname{equation}{}{}
\usepackage[per-mode=symbol]{siunitx}
\usepackage{tikz}
\usepackage{pgfplots}\pgfplotsset{compat=1.16}\newlength{\figurewidth}\newlength{\figureheight}
\usepgfplotslibrary{groupplots}
\usepackage{tabularx}
\usepackage{booktabs}\newcommand{\otoprule}{\midrule[\heavyrulewidth]}
\usepackage{subcaption}

\usepackage[utf8]{inputenc}   				 	
\usepackage{url}
\usepackage{natbib}
\usepackage{letltxmacro}
\LetLtxMacro{\autocite}{\citep}
\LetLtxMacro{\textcite}{\citet}

\makeatletter
\newcommand*{\org@overidelabel}{}
\let\org@overridelabel\@verridelabel
\renewcommand*{\@verridelabel}[1]{%
  \@bsphack
  \protected@write\@auxout{}{\string\AC@undonewlabel{#1@cref}}%
  \org@overridelabel{#1}%
  \@esphack
}%
\makeatother

\title{PRISMA: A Novel Approach for Deriving Probabilistic Surrogate Safety Measures for Risk Evaluation}
\author[1,2]{Erwin de Gelder\corref{cor1}}
\ead{erwin.degelder@tno.nl}
\author[3]{Kingsley Adjenughwure}
\author[1]{Jeroen Manders}
\author[4]{Ron Snijders}
\author[1,5]{Jan-Pieter Paardekooper}
\author[1]{Olaf Op den Camp}
\author[1,6]{Arturo Tejada}
\author[2]{Bart De Schutter}
\cortext[cor1]{Corresponding author}
\address[1]{TNO, Integrated Vehicle Safety, Helmond, The Netherlands}
\address[2]{Delft University of Technology, Delft Center for Systems and Control, Delft, The Netherlands}
\address[3]{TNO, Sustainable Urban Mobility \& Safety, The Hague, The Netherlands}
\address[4]{TNO, Monitoring \& Control Services, Groningen, The Netherlands}
\address[5]{Radboud University, Donders Institute for Brain, Cognition and Behaviour, Nijmegen, The Netherlands}
\address[6]{Eindhoven University of Technology, Dynamics and Control Group, Eindhoven, The Netherlands}
\date{}

\let\originalleft\left
\let\originalright\right
\renewcommand{\left}{\mathopen{}\mathclose\bgroup\originalleft}
\renewcommand{\right}{\aftergroup\egroup\originalright}
\newcommand{\accelerationsymbol}{a}
  \newcommand{\accelerationmax}{\accelerationsymbol_{\mathrm{MADR}}}
  \newcommand{\accelerationleadsymbol}{\accelerationsymbol_{\mathrm{l}}}
  \newcommand{\accelerationlead}[1]{\accelerationleadsymbol\left(#1\right)}
\newcommand{\bandwidth}{h}
\newcommand{\bandwidthmatrix}{H}
  \newcommand{\bandwidthnw}{H_{\mathrm{NW}}}
\newcommand{\collision}{\mathrm{C}}
\newcommand{\density}[1]{p\left( #1 \right)}
  \newcommand{\densitycond}[2]{\density{ #1 | #2 }}
  \newcommand{\densityest}[1]{\hat{p}\left( #1 \right)}
  \newcommand{\densityestcond}[2]{\densityest{ #1 | #2 }}
\newcommand{\dimension}{d}

  \newcommand{\dummyvara}{a}
  \newcommand{\dummyvarb}{b}
  \newcommand{\dummyvarkernel}{u}
\newcommand{\dimsimulationresult}{n_{\mathrm{z}}}
\newcommand{\e}[1]{\exp \left\{ #1 \right\}}

\newcommand{\gapsymbol}{g}
  \newcommand{\gap}[1]{\gapsymbol\left(#1\right)}
\newcommand{\identitymatrix}[1]{I_{#1}}
\newcommand{\kernelfuncnormalized}[2]{K_{#1} \left( #2 \right)}
\newcommand{\lowerbound}{L}
  \newcommand{\lowerboundadapted}{\hat{\lowerbound}}
\newcommand{\lnsymbol}{\log}

\newcommand{\normtwo}[1]{\left\Vert #1 \right\Vert_2}
\newcommand{\numberofcollisions}{N_{\collision}}
  \newcommand{\numberofdesignpoints}{m}
  
  \newcommand{\numberofsimulations}{N_{\mathrm{sim}}}
  \newcommand{\numberoftrafficparticipants}{N_{\mathrm{tp}}}
\newcommand{\probability}[1]{P\left( #1 \right)}
  \newcommand{\probabilitycond}[2]{\probability{ #1 | #2 }}
  \newcommand{\probabilityest}[1]{\hat{P}\left( #1 \right)}
  \newcommand{\probabilityestcond}[2]{\probabilityest{ #1 | #2 }}
\newcommand{\realnumbers}{\mathds{R}}
\newcommand{\spacecollision}{\mathcal{Z}_{\collision}}
\newcommand{\speeddifferencesymbol}{\Delta v}
  \newcommand{\speeddifference}[1]{\speeddifferencesymbol \left( #1 \right)}
\newcommand{\simulationindex}{j}
\newcommand{\simulationresult}{z}
  \newcommand{\simulationinstance}[1]{\simulationresult_{#1}}
  \newcommand{\simulationbandwidth}{\bandwidthmatrix_{\simulationresult}}
\newcommand{\simulationthreshold}{\epsilon}
\newcommand{\situationinitial}{x}
  \newcommand{\situationinitialdim}{n_{\situationinitial}}
  \newcommand{\situationinitialinstance}[1]{x_{#1}}
  \newcommand{\situationinitialmean}{\mu_{x}}
  \newcommand{\situationinitialspace}{\mathcal{X}}
  \newcommand{\situationinitialtp}[1]{x^{#1}}
\newcommand{\situationfuture}{y}
  \newcommand{\situationfuturedim}{n_{\situationfuture}}
  \newcommand{\situationfuturehorizon}{n_{\mathrm{h}}}
  \newcommand{\situationfuturetimestep}{\Delta \time}
  \newcommand{\situationfutureinstance}[1]{y_{#1}}
  \newcommand{\situationfuturemean}{\mu_{y}}
  \newcommand{\situationfutureparta}{\bar{\situationfuture}}
  \newcommand{\situationfuturepartb}{\tilde{\situationfuture}}
  \newcommand{\situationfuturespace}{\mathcal{Y}}
\newcommand{\situationindex}{i}
  \newcommand{\situationindexdesign}{k}
\newcommand{\situationnumberof}{N}
\newcommand{\speedsymbol}{v}
  \newcommand{\speedegosymbol}{\speedsymbol_{\mathrm{e}}}
  \newcommand{\speedego}[1]{\speedegosymbol\left(#1\right)}
  \newcommand{\speedleadsymbol}{\speedsymbol_{\mathrm{l}}}
  \newcommand{\speedlead}[1]{\speedleadsymbol\left(#1\right)}
\newcommand{\svdu}{U}
  \newcommand{\svduupperleft}{\bar{\svdu}_1}
  \newcommand{\svdulowerleft}{\bar{\svdu}_2}
  \newcommand{\svduvec}[1]{u_{#1}}
  \newcommand{\svds}{\Sigma}
  \newcommand{\svdsv}[1]{\sigma_{#1}}
  \newcommand{\svdsupperleft}{\bar{\svds}}
  \newcommand{\svdv}{V}
  \newcommand{\svdventry}[2]{v_{#1 #2}}
  \newcommand{\svdvvecsymbol}{\bar{v}}
  \newcommand{\svdvvecd}[1]{\svdvvecsymbol_{#1}}
  \newcommand{\svdrank}{\bar{N}}
  \newcommand{\svdindex}{j}
\renewcommand{\time}{t}
  \newcommand{\timeend}{t_{\mathrm{end}}}
  \newcommand{\timereact}{\time_{\mathrm{r}}}
  \newcommand{\timemaxreact}{\time_{\mathrm{max}}}
\newcommand{\trafficparticipantindex}{i}
  \newcommand{\trafficparticipantindexb}{j}
\newcommand{\transpose}{^{\mkern-1.5mu\mathsf{T}}}
\newcommand{\ttcsymbol}{t_\mathrm{TTC}}
\newcommand{\ttc}[1]{\ttcsymbol\left( #1 \right)}
\newcommand{\ud}{\,\mathrm{d}}
\newcommand{\upperbound}{U}
\newcommand{\variance}[1]{\mathrm{Var}\left[#1\right]}
\newcommand{\wangstamatiadis}{\mathrm{WS}}
\newcommand{\weightmatrix}{W}

\newcommand{\cstart}{}
\newcommand{\cend}{}

\begin{document}

\begin{acronym}[AAAAAAAA]
	\acro{ads}[ADS]{Automated Driving System}\acroindefinite{ads}{an}{an}
	\acro{cpi}[CPI]{Crash Potential Index}
	\acro{drac}[DRAC]{Deceleration Rate to Avoid Collision}
	\acro{evt}[EVT]{Extreme Value Theory}
	\acro{kde}[KDE]{Kernel Density Estimation}
	\acro{idm}[IDM]{Intelligent Driver Model}
	\acro{idmplus}[IDM+]{Intelligent Driver Model Plus}
	\acro{madr}[MADR]{Maximum Available Deceleration Rate}
	\acro{mttc}[MTTC]{Modified TTC}
	\acro{ngsim}[NGSIM]{Next Generation SIMulation}
	\acro{nw}[NW]{Nadaraya-Watson}\acroindefinite{nw}{an}{a}
	\acro{odd}[ODD]{Operational Design Domain}\acroindefinite{odd}{an}{an}
	\acro{ourmethod}[PRISMA]{Probabilistic RISk Measure derivAtion}
	\acro{pca}[PCA]{Principal Component Analysis}
	\acro{pdf}[PDF]{probability density function}
	\acro{pet}[PET]{Post-Encroachment Time}
	\acro{picud}[PICUD]{Potential Index for Collision with Urgent Deceleration}
	\acro{psd}[PSD]{Proportion of Stopping Distance}
	\acro{ssm}[SSM]{Surrogate Safety Measure}\acroindefinite{ssm}{an}{a}
	\acro{svd}[SVD]{Singular Value Decomposition}\acroindefinite{svd}{an}{a}
	\acro{thw}[THW]{Time Headway}
	\acro{tit}[TIT]{Time Integrated TTC}
	\acro{ttc}[TTC]{Time to Collision}
	\acro{ttcd}[TTCD]{Time to Collision Disturbance}
\end{acronym}

\begin{abstract}

	\acp{ssm} are used to express road safety in terms of the safety risk in traffic conflicts.
	Typically, \acp{ssm} rely on assumptions regarding the future evolution of traffic participant trajectories to generate a measure of risk, \cstart restricting their applicability to scenarios where these assumptions are valid. \cend
	\cstart In response to this limitation, we present the novel \cend\ac{ourmethod} method.
	\cstart The objective of \cend the \ac{ourmethod} method is to derive \acp{ssm} that can be used to calculate in real time the probability of a specific event (e.g., a crash). 
	\cstart The \ac{ourmethod} method adopts \cend a data-driven approach to predict the possible future traffic participant trajectories, \cstart thereby reducing the reliance on specific assumptions regarding these trajectories. 
	Since the \ac{ourmethod} is not bound to specific assumptions, \cstart the \ac{ourmethod} method offers the ability to derive \cend multiple \acp{ssm} for \cstart various \cend scenarios. 
	\cstart The occurrence probability of the specified event is based on simulations and combined \cend with a regression model, this enables our derived \acp{ssm} to make real-time risk estimations.
	
	To illustrate the \ac{ourmethod} method, \iac{ssm} is derived for risk evaluation during longitudinal traffic interactions. 
	Since there is no known method to objectively estimate risk from first principles, i.e., there is no known risk ground truth, it is very difficult, if not impossible, to objectively compare the relative merits of two \acp{ssm}.
	Instead, we provide a method for benchmarking our derived \ac{ssm} with respect to expected risk trends.
	The application of the benchmarking illustrates that the \ac{ssm} matches the expected risk trends.
	
	Whereas the derived \ac{ssm} shows the potential of the \ac{ourmethod} method, future work involves applying the approach for other types of traffic conflicts, such as lateral traffic conflicts or interactions with vulnerable road users.

\end{abstract}

\maketitle
\acresetall
\section{Introduction}
\label{sec:introduction}

Road safety is an important key performance indicator in transportation. 
In addition to the suffering of people as a consequence of crashes in traffic, these crashes cause enormous societal and economic losses.
As a result, road safety research is an important research topic.
For example, in 2018\footnote{At the time of writing, more recent results were not yet available.}, there were over 6.7 million crashes in the U.S.A.\ \autocite{nhtsa2020summary}, which is about 1.3 crashes per 1 million vehicle kilometers driven.
These crashes in 2018 led to 2.7 million injured people and 37 thousand fatalities \autocite{nhtsa2020summary}.
Furthermore, apart from these societal losses, the economic costs of all crashes in the U.S.A.\ in 2018 was 242 billion dollars \autocite{nhtsa2020summary}.
Similarly, the \textcite{eu2020roadsafety} reported over 22 thousand fatalities in 2019.

Road safety can be expressed in terms of injuries, fatalities, or crashes per kilometer of driving, but ``that is a slow, reactive process'' \autocite{arun2021systematic}.
Furthermore, ``crashes are rare events and historical crash data does not capture near crashes that are also critical for improving safety'' \autocite{wang2021review}.
An alternative for expressing road safety that does not rely on historical crash data is the use of safety indicators that directly measure the safety risk in traffic conflicts \autocite{tarko2018surrogate, arun2021systematic, wang2021review}.
Traffic conflicts are far more frequent than traffic crashes and the frequency of traffic conflicts can be used to predict the frequency of crashes \autocite{davis2011outline, tarko2018estimating}.
To define traffic conflicts, thresholds on so-called \acp{ssm} are used, where \acp{ssm} characterize the risk of a crash or harm given an initial condition \autocite{arun2021systematic}.
\acp{ssm} vary from measures that estimate the remaining time until a crash, such as the well-known \ac{ttc} \autocite{hayward1972near}, to metrics that estimate the probability that a human driver cannot avoid a crash, see, e.g., \autocite{wang2014evaluation}.

\Acp{ssm} typically rely on assumptions of what drivers or systems controlling the vehicles of interest are capable of doing and how their future trajectories --- given an initial condition --- will develop. 
For example, \ac{ttc} \autocite{hayward1972near}, the ratio of the distance toward and the speed difference with an approaching object, is computed by assuming a constant relative velocity. 
As a result of these assumptions, \acp{ssm} are only applicable in certain types of scenarios.
For example, \ac{ttc} is only applicable when approaching an object.
More complex \acp{ssm} consider, e.g., a human model that can react to a risky situation by braking \autocite{wang2014evaluation} or the uncertainty over the future ambient traffic state \autocite{mullakkal2020probabilistic}.
Regardless of the complexity of these models, however, these \acp{ssm} consider neither the specific capabilities of the driver or of the system controlling the vehicle, nor the local context for predicting the future of the vehicle's environment.  

This paper presents the \ac{ourmethod} method, which is a data-driven approach for deriving \acp{ssm} that are not limited to certain types of scenarios.
Because the method is not bound to certain predetermined assumptions about driver behavior, the derived \acp{ssm} can be adapted to the situations in which they are applied. 
In addition, to avoid relying on predetermined assumptions on how the ambient traffic evolves over time, the \ac{ourmethod} method includes a data-driven approach for modeling the variations of the trajectories of the ambient traffic.
Monte Carlo simulations are employed to predict the safety risk given these variations.
To enable the real-time evaluation of the derived \acp{ssm}, we use the \ac{nw} kernel estimator \autocite{wasserman2006nonparametric} for local regression.
The \ac{ourmethod} method has the following characteristics:
\begin{itemize}
	\item The derived \acp{ssm} give a probability that a specified event, e.g., a crash or a near miss will happen in the near future, e.g., within the next 10 seconds, given an initial state and the foreseen evolutions of traffic participant trajectories. 
	Since a traffic conflict can be defined as the probability of an unsuccessful evasion in a traffic interaction, according to \textcite{davis2011outline}, a probability is easier to interpret than, e.g., a value ranging from 0 to infinity such as the \ac{ttc}.
	
	\item Next to deriving new \acp{ssm}, i.e., new ways to estimate the probability of an event such as a crash, it is possible to reproduce already existing measures that provide a probability. 
	Therefore, the \ac{ourmethod} method can be seen as a generalization for deriving such existing \acp{ssm}.
	
	\item A driver behavior model can be used.
	It is also possible to use a model of \iac{ads}, such that the derived \ac{ssm} estimates the safety risk if this \ac{ads} controls the vehicle.
	
	\item Because a data-driven approach is adopted, the derived \ac{ssm} adapts to the recorded data. 
	In this way, it is possible to adapt the \ac{ssm} to, e.g., the local traffic behavior provided that this local traffic behavior is captured by the recorded data.
	
	\item The \ac{ourmethod} method is not limited to one type of scenario.
\end{itemize}

We illustrate the \ac{ourmethod} method and its benefits by means of a case study. 
The case study demonstrates that when using the \ac{ourmethod} method with the assumptions of the \ac{ssm} of \textcite{wang2014evaluation}, both the \ac{ssm} derived by the \ac{ourmethod} method and the latter yield the same result.
The case study continues with evaluating the crash risk of three longitudinal traffic conflicts which are a priori known to be, respectively, safe (i.e., no crash possible), moderately safe, and unsafe (i.e., a crash occurs), based on vehicle kinematics. 
We evaluate the risk of each of the scenarios using the \ac{ssm} by \textcite{wang2014evaluation} and \iac{ssm} derived by the \ac{ourmethod} method, based on data from the \ac{ngsim} \autocite{alexiadis2004next}. 
Moreover, since a comparison between these measures is not directly possible in general scenarios, a method to benchmark \acp{ssm} using expected risk trends is introduced in the case study.

This article is organized as follows.
\cref{sec:literature review} provides an overview of \acp{ssm} described in the literature.
The proposed \ac{ourmethod} method is presented in \cref{sec:method}.
In \cref{sec:case study}, we illustrate the method in a case study.
Some implications of this work are discussed in \cref{sec:discussion}.
The article is concluded in \cref{sec:conclusions}.

\section{Literature review}
\label{sec:literature review}

Risk in the context of traffic safety is often defined as the probability of a crash occurring \autocite{hakkert2002uses}.
Most \acp{ssm} are derived under specific assumptions of the expected behavior of the  driving participants under a specific  driving scenario.
Several \acp{ssm} have been developed under such assumptions with the goal of quantifying the risk involved in driving in traffic on the road \autocite{minderhoud2001extended, ozbay2008derivation, cunto2009simulated, laureshyn2010evaluation}.
In general, the risk is quantified in terms of the proximity between two traffic agents in time and/or space, the ability to perform evasive actions like braking or swerving, or the magnitude of such actions \autocite{shi2018key,zheng2020modeling}. 
In a potential crash situation, the proximity indicator is close to zero while the magnitude of evasive action is close to the limits of the driver and the vehicle \autocite{zheng2020modeling}. 
The above clustering of \acp{ssm} in terms of time, space, and evasive action is common in the literature; so our literature review follows this pattern of clustering \acp{ssm}.
We focus on the most commonly used measures in each cluster and their underlying assumptions.
\cstart Additionally, we discuss some well-known \ac{ssm}-based metrics, i.e., metrics that are derived from other \acp{ssm}. \cend

The most common \acp{ssm} are time-based. 
A popular time-based proximity indicator is the \ac{ttc}, which is an estimate of the remaining time until two vehicles collide and is defined as the time remaining until two vehicles collide if they would continue on the same course and speed \autocite{hayward1972near}.
The assumption for the \ac{ttc} is that the relative speed and course will remain the same. 
In addition, the \ac{ttc} is only relevant when two objects are approaching each other. 
These assumptions make it difficult to use it  for various driving scenarios.
Several other time-based \acp{ssm} have been derived from or based on the \ac{ttc}. 
Notable among those are:
\begin{itemize}
    \item the time-exposed \ac{ttc}, which measures the amount of time the \ac{ttc} is below a certain threshold \autocite{minderhoud2001extended};
    \item the \ac{tit}, which calculates the total area in \iac{ttc} versus time diagram where the \ac{ttc} is below a certain threshold  \autocite{minderhoud2001extended};
    \item the \ac{mttc}, which is able to calculate the \ac{ttc} for cases where vehicles do not keep a constant speed and the follower is slower than the leader \autocite{ozbay2008derivation}; \cstart and
    \item the \ac{ttcd}, which calculates the \ac{ttc} in case the leader is decelerating with a constant deceleration \autocite{xie2019use}. \cend
\end{itemize}
For the \ac{mttc} \cstart and \ac{ttcd}\cend, the relative speed is not assumed to be constant, but new assumptions on the acceleration and speed of the objects are introduced. 

Other time-based proximity indicators include \ac{pet}, which measures the ``time between the moment that a vehicle leaves the area of potential collision, i.e., the area in which the paths of the two vehicles intersect, and the other vehicle arrives in the same area'' \autocite{mahmud2017application} and \ac{thw}.
\ac{pet} can only be calculated when the collision area of the two participants is known. 
This assumption makes it mostly useful for scenarios with obvious crossing conflicts like intersections.

For distance-based proximity indicators, the \ac{picud} measures the remaining distance between vehicles during an emergency stop \autocite{iida2001traffic, uno2003objective} and the \ac{psd} measures the remaining distance to the potential point of collision divided by the minimum acceptable stopping distance \autocite{allen1978analysis, guido2011comparing, mahmud2017application}. 
These two measures assume that the vehicles will apply the maximum deceleration during emergency situations. 
This makes them suitable for emergency situations for which these assumption will most likely hold. 
For non-critical situations, however, the deceleration that the drivers will apply, may vary.
More recently, a distance-based measure that assumes ``correct'' driving behavior has been proposed \autocite{shalev2017formal}. 
This measure calculates the minimum safety distance between a follower and its leader, such that no crash occurs if the leader vehicle brakes with a specified deceleration and the follower brakes after a specified reaction time with another specified deceleration. 
Based on the definition of this measure, it is not suitable for driving situations where the driver does not follow the description of ``correct'' driving given above.

In terms of indicators relating to performing evasive actions, the \ac{drac} is the most widely used. 
The \ac{drac} is calculated as the ratio of the difference in speed between a following vehicle and a leading vehicle and their closing time \autocite{almqvist1991use, mahmud2017application}. 
Another indicator is the \ac{cpi}, which calculates the probability that a vehicle's \ac{drac} will exceed its \ac{madr} in a given time interval \autocite{cunto2009simulated}. 
The \ac{drac} is not a risk measure on its own, if it is not compared with the braking capacity.
This is a limitation and this is why the \ac{cpi} measure has been developed.
Both \ac{drac} and \ac{cpi} are mostly suitable for a car-following situation and are not suitable for lateral movements \autocite{mahmud2017application}.

\cstart\textcite{wang2014evaluation} and \textcite{xie2019use} propose \iac{ssm}-based metric, i.e., they derive a probability using the \ac{ttc} and \ac{ttcd}, respectively, and some assumptions. 
In particular, \textcite{wang2014evaluation} assume distributions of the vehicle braking capability and the driver's reaction time, while \textcite{xie2019use} assume a distribution of the deceleration rate of the leader.
Although these probabilities are suitable for various car-following situations, lane-change conflicts, and crossing conflicts, they are limited because they use the \ac{ttc} and \ac{ttcd}, respectively, in their calculations; so these probabilities are undefined when the \ac{ttc} and \ac{ttcd}, respectively, are undefined.

\textcite{saunier2008probabilistic} derive a probability based on the \acp{ttc} that are estimated using hypothetical trajectories of the different traffic participants. 
Similar to the approach we present in \cref{sec:method}, the hypothetical trajectories of the traffic participants are based on a data-driven model \autocite{saunier2007probabilistic}. 
In \autocite{saunier2008probabilistic}, also the probabilities of the different hypothetical trajectories are considered. 
In order to keep the computation of the metric feasible in real time, the number of hypothetical trajectories to be considered is limited.
\textcite{altendorfer2021new} also calculate the probability of a collision. 
They provide a general framework that can consider any arbitrary prediction model of the trajectories of the traffic participants.
To ensure real-time computations, their example considers Gaussian distributions and simplified geometries of the traffic participants, such that they can use numerical integration instead of Monte Carlo simulations. \cend

\textcite{shi2018key} use indicators like \ac{tit}, \ac{cpi}, and \ac{psd} to measure the effectiveness of risk indicators for predicting crashes. 
The idea is to use a combination of indicators and thresholds on the indicators to predict whether an interaction may become a crash.
This results in new indicators, but they inherit the union of the assumptions of the other indicators.
\textcite{mullakkal2020probabilistic} propose a probabilistic driving risk field.
The method derives the risk a vehicle is exposed to using a kinematic approach with the inclusion of uncertainty in the vehicle's future state. 
\textcite{mullakkal2020probabilistic} define this for an encounter between the ego vehicle and a road obstacle, such as other vehicles or objects. 
This research shares similar ideas with our proposed method of risk estimation, but \textcite{mullakkal2020probabilistic} do not use a data-driven approach to derive the \ac{ssm}. 
Furthermore, the future state of the vehicle is estimated with a fixed distribution (i.e., a normal distribution). 
This limits the application in scenarios where the data may have an entirely different distribution.

To estimate crash probabilities based on existing \acp{ssm}, the probabilistic approach using the \ac{evt} has been applied successfully \autocite{songchitruksa2006extreme, tarko2012use, wang2021review}.
For example, based on a specific \ac{ttc} value, \ac{evt} can be used to predict the probability of a crash. 
Using \ac{evt}, the crash probability is estimated by assuming the generalized extreme value distribution and fitting the parameters of the distribution using either the ``block maxima'' approach or the ``peak over'' approach \autocite{wang2021review}.
It is also possible to combine multiple \acp{ssm} using the \ac{evt}.
The advantage of \ac{evt} is that \ac{evt} provides probabilities that are directly linked to historical data and that these probabilities have been used successfully to predict the frequency of crashes \autocite{songchitruksa2006extreme, aasljung2017using}.
Disadvantages of \ac{evt} are that it might inherit the assumptions of the \acp{ssm} that it uses to estimate the crash probability and that it assumes a fixed distribution of the extreme events, which is only justified if a lot of data is used.
Furthermore, as the estimated crash probability is solely based on the fitted distribution, it does not consider potential changes to the driver's behavior (model).

\section{Probabilistic RISk Measure derivAtion}
\label{sec:method}

In this section, we propose the \ac{ourmethod} method which is a method for deriving a measure that quantifies the risk of a certain event, such as a crash, in a particular situation in which a vehicle - hereafter, the \textit{ego vehicle} - is in and that is applicable for real-time use.
The \ac{ourmethod} method \cstart is schematically shown in \cref{fig:outline} and \cend consists of four steps:
\cstart
\begin{enumerate}
	\item The parameterization of the ``initial situation'' and the possible ``future situations'' (\cref{sec:parametrization});
	\item Based on the initial situation, the estimation of the probability (density) for the possible future situations (\cref{sec:estimate future});
	\item The estimation of the probability of the specified event based on the initial and the future situations (\cref{sec:estimate collision}); and
	\item Local regression in order to speed up the calculations and to make it possible to use the \ac{ssm} in real time (\cref{sec:final metric calculation}). 
\end{enumerate}\cend

\begin{figure}
	\centering
	\newlength{\blockwidth}\setlength{\blockwidth}{8em}%
\newlength{\blockheight}\setlength{\blockheight}{6.5em}%
\newlength{\blocksepx}\setlength{\blocksepx}{1.5em}%
\newlength{\blocksepy}\setlength{\blocksepy}{1em}%
\newlength{\legendheight}\setlength{\legendheight}{1.4em}%
\newlength{\legendsep}\setlength{\legendsep}{0.3em}%
\tikzstyle{block}=[draw, text width=\blockwidth-.5em, align=center, minimum height=\blockheight, very thick, minimum width=\blockwidth, anchor=north west, rounded corners=0.2em]%
\tikzstyle{blockwide}=[block, text width=2\blockwidth+\blocksepx-.5em, minimum width=2\blockwidth+\blocksepx]%
\tikzstyle{blocklegend}=[block, minimum height=\legendheight]%
\tikzstyle{sec1}=[draw=blue, fill=blue!20]%
\tikzstyle{sec2}=[draw=green, fill=green!20, dashed]%
\tikzstyle{sec3}=[draw=yellow, fill=yellow!20, dotted]%
\tikzstyle{sec4}=[draw=red, fill=red!20, dash dot]%
\tikzstyle{arrow}=[->, very thick]%
\begin{tikzpicture}
	\node[block, sec1](parameters) {Parameterization \\ initial situation ($\situationinitial$) and future situation ($\situationfuture$)};
	\node[block, sec2](data) at (\blockwidth+\blocksepx, 0) {Data with initial and future situations: $\left\{(\situationinitialinstance{\situationindex},\situationfutureinstance{\situationindex})\right\}_{\situationindex=1}^{\situationnumberof}$};
	\node[blockwide, sec2](pdf) at (0, -\blockheight-\blocksepy) {Estimate probability density of future situation given initial situation: $\densitycond{\situationfuture}{\situationinitial}$};
	\node[block, sec3](mc) at (2\blockwidth+2\blocksepx, 0) {Method for estimating event probability given initial situation: $\probabilitycond{\collision}{\situationinitial}$};
	\node[block, sec4](design points) at (3\blockwidth+3\blocksepx, 0) {Initial situations for conducting the simulations: $\left\{\situationinitialinstance{\situationindexdesign}'\right\}_{\situationindexdesign=1}^{\numberofdesignpoints}$};
	\node[blockwide, sec4](doing mc) at (2\blockwidth+2\blocksepx, -\blockheight-\blocksepy) {Estimate event probability for each initial situation of the set $\left\{\situationinitialinstance{\situationindexdesign}'\right\}_{\situationindexdesign=1}^{\numberofdesignpoints}$: \\ $\probabilityestcond{\collision}{\situationinitialinstance{\situationindexdesign}'}$ for all $\situationindexdesign$ in $\left\{1,\ldots,m\right\}$};
	\node[block, sec4](end) at (4\blockwidth+4\blocksepx, -\blockheight-\blocksepy) {Use regression to estimate event probability for any $\situationinitial$: $\probabilityestcond{\collision}{\situationinitial}$};
	
	\draw[arrow](\blockwidth/2, -\blockheight) -- (\blockwidth/2, -\blockheight-\blocksepy);
	\draw[arrow](3\blockwidth/2+\blocksepx, -\blockheight) -- (3\blockwidth/2+\blocksepx, -\blockheight-\blocksepy);
	\draw[arrow](pdf) -- (doing mc);
	\draw[arrow](5\blockwidth/2+2\blocksepx, -\blockheight) -- (5\blockwidth/2+2\blocksepx, -\blockheight-\blocksepy);
	\draw[arrow](7\blockwidth/2+3\blocksepx, -\blockheight) -- (7\blockwidth/2+3\blocksepx, -\blockheight-\blocksepy);
	\draw[arrow](doing mc) -- (end);
	
	\node[blocklegend, sec1] at (4\blockwidth+4\blocksepx, 0) {\cref{sec:parametrization}};
	\node[blocklegend, sec2] at (4\blockwidth+4\blocksepx, -\legendheight-\legendsep) {\cref{sec:estimate future}};
	\node[blocklegend, sec3] at (4\blockwidth+4\blocksepx, -2\legendheight-2\legendsep) {\cref{sec:estimate collision}};
	\node[blocklegend, sec4] at (4\blockwidth+4\blocksepx, -3\legendheight-3\legendsep) {\cref{{sec:final metric calculation}}};	
\end{tikzpicture}
	\caption{\cstart Schematic overview of the \ac{ourmethod} method and the organization of \cref{sec:method}.
		The mathematical symbols are further explained in \cref{sec:parametrization,sec:estimate future,sec:estimate collision,sec:final metric calculation}.\cend}
	\label{fig:outline}
\end{figure}

In this article, the following notation is used. 
To denote a probability function, $\probability{\cdot}$ is used. 
\Iac{pdf} is denoted by $\density{\cdot}$. 
The probability of $\dummyvara$ given $\dummyvarb$ is denoted by $\probabilitycond{\dummyvara}{\dummyvarb}$.
Similarly, a conditional \ac{pdf} is denoted by $\densitycond{\cdot}{\cdot}$. 
To denote the estimation of any of the aforementioned functions, a circumflex is used, e.g, $\probabilityest{\dummyvara}$ denotes the estimated probability of $\dummyvara$.

\subsection{Parameterize initial and future situations}
\label{sec:parametrization}

The first step is to parameterize the initial situation the ego vehicle is in. 
In other words, the initial situation needs to be described using $\situationinitialdim$ numbers that are stacked into one vector $\situationinitial \in \situationinitialspace \subseteq \realnumbers^{\situationinitialdim}$. 
This vector contains relevant aspects for determining the risk.
As an example, $\situationinitial$ could contain the speed of the ego vehicle and the distance toward its preceding vehicle. 
In \cref{sec:case study}, we will consider more examples.

Next to describing the initial situation, the future situation is described using $\situationfuturedim$ numbers stacked into one vector $\situationfuture \in \situationfuturespace \subseteq \realnumbers^{\situationfuturedim}$. 
Together with $\situationinitial$, $\situationfuture$ contains enough information to describe how the relevant future, e.g., the next 5 seconds, around the ego vehicle develops over time. 
As an example, $\situationfuture$ could contain the speed for the next 5 seconds of the leading vehicle (if any) that is in front of the ego vehicle.
In \cref{sec:case study}, we will consider more examples.

Let $\collision$ denote an event, e.g., a crash or a near miss, such that the probability of this event is $\probability{\collision}$.
The goal of our \ac{ssm} is to estimate the probability of the event $\collision$ given a particular situation $\situationinitial$, i.e., $\probabilitycond{\collision}{\situationinitial}$.
We do this by considering all future situations, $\situationfuturespace$, and calculating the probability of the event $\collision$ given each possible value of $\situationfuture$. 
Using integration, we obtain $\probabilitycond{\collision}{\situationinitial}$:
\begin{equation}
	\label{eq:probability collision expectation}
	\probabilitycond{\collision}{\situationinitial} 
	= \int_{\situationfuturespace} 
	\probabilitycond{\collision}{\situationinitial, \situationfuture} 
	\densitycond{\situationfuture}{\situationinitial} 
	\ud \situationfuture.
\end{equation}

\subsection{Estimate $\densitycond{\situationfuture}{\situationinitial}$}
\label{sec:estimate future}

In this section, we propose a method to estimate $\densitycond{\situationfuture}{\situationinitial}$, i.e., the \ac{pdf} of $\situationfuture$ given $\situationinitial$.
Using the product rule for probability, we can write:
\begin{equation}
	\densitycond{\situationfuture}{\situationinitial} 
	= \frac{\density{\situationinitial, \situationfuture}}{\density{\situationinitial}}
	= \frac{\density{\situationinitial, \situationfuture}}{
		\int_{\situationfuturespace} \density{\situationinitial, \situationfuture} \ud\situationfuture
	}.
\end{equation}
Thus, it suffices to estimate $\density{\situationinitial, \situationfuture}$. 

Our proposal is to estimate $\density{\situationinitial, \situationfuture}$ in a data-driven manner. 
A data-driven approach brings several benefits.
First, the estimate automatically adapts to local driving styles and behaviors, which can change from region to region, provided that the data are obtained from the same local traffic.
Second, assumptions such as a constant speed of other vehicles, are not needed.
For our data-driven approach, let us assume that we have obtained $\situationnumberof$ situations from data.
For the $\situationindex$-th situation, we denote the initial situation and the future situation by $\situationinitialinstance{\situationindex}\in\situationinitialspace$ and $\situationfutureinstance{\situationindex}\in\situationfuturespace$, respectively. 
The remainder of this subsection describes how we estimate $\density{\situationinitial, \situationfuture}$ using  $\left\{(\situationinitialinstance{\situationindex},\situationfutureinstance{\situationindex})\right\}_{\situationindex=1}^{\situationnumberof}$.

\subsubsection{Kernel density estimation}
\label{sec:one kde}

We first explain how to estimate $\density{\situationinitial, \situationfuture}$ if we assume that all $\situationinitialdim+\situationfuturedim$ parameters depend on each other. 
If the shape of the \ac{pdf} is known, a particular functional form can be fitted to the data, e.g., by estimating the parameters of a distribution by maximizing the likelihood.
For example, if it is known that the data $\left\{(\situationinitialinstance{\situationindex},\situationfutureinstance{\situationindex})\right\}_{\situationindex=1}^{\situationnumberof}$ come from a multivariate normal distribution, it suffices to estimate the mean and the covariance.
If, however, the shape is unknown, fitting a particular parametric distribution may lead to very inaccurate results \autocite{chen2017tutorial}.
Furthermore, the shape of the estimated \ac{pdf} might change as more data are acquired. 
Assuming a functional form of the \ac{pdf} and fitting the parameters of the \ac{pdf} to the data may therefore lead to inaccurate fits unless extensive manual tuning is applied.

In the remainder of this work, we assume that the shape of the \ac{pdf} $\density{\situationinitial, \situationfuture}$ is unknown a priori.
Therefore, we employ a non-parametric approach using \ac{kde} \autocite{rosenblatt1956remarks, parzen1962estimation} because the shape of the \ac{pdf} is then automatically computed and \ac{kde} is highly flexible regarding the shape of the \ac{pdf}. 
\cstart Note, however, that the \ac{ourmethod} method can also work with other non-parametric methods for estimating \iac{pdf} (cf.\ \autocite{durkan2019neural, peerlings2022multivariate}). \cend
Using \ac{kde}, the estimated \ac{pdf} becomes:
\begin{equation}
	\label{eq:kde estimate}
	\densityest{\situationinitial,\situationfuture}
	= \frac{1}{\situationnumberof} \sum_{\situationindex=1}^{\situationnumberof}
	\kernelfuncnormalized{\bandwidthmatrix}{
		\begin{bmatrix}
			\situationinitial \\
			\situationfuture
		\end{bmatrix} -
		\begin{bmatrix}
			\situationinitialinstance{\situationindex} \\
			\situationfutureinstance{\situationindex}
		\end{bmatrix}
	},
\end{equation}
where $\kernelfuncnormalized{\bandwidthmatrix}{\cdot}$ is an appropriate kernel function with an $(\situationinitialdim+\situationfuturedim)$-by-$(\situationinitialdim+\situationfuturedim)$ symmetric positive definite \emph{bandwidth} or \emph{smoothing} matrix $\bandwidthmatrix$. 
The choice of the kernel $\kernelfuncnormalized{\bandwidthmatrix}{\cdot}$ is not as important as the choice of the bandwidth matrix $\bandwidthmatrix$ \autocite{turlach1993bandwidthselection}.
We use the often-used Gaussian kernel \autocite{duong2007ks}:
\begin{equation}
	\label{eq:kernel initial future}
	\kernelfuncnormalized{\bandwidthmatrix}{\dummyvarkernel}
	= \frac{1}{\left( 2 \pi \right)^{\left( \situationinitialdim + \situationfuturedim \right) / 2} 
	\left|\bandwidthmatrix\right|^{1/2} }
	\e{ -\frac{1}{2} \dummyvarkernel\transpose \bandwidthmatrix^{-1} \dummyvarkernel }.
\end{equation}

The bandwidth matrix $\bandwidthmatrix$ controls the width of the kernel, or, in other words, the influence of each data point (i.e., $\begin{bmatrix}\situationinitialinstance{\situationindex}\transpose & \situationfutureinstance{\situationindex}\transpose\end{bmatrix}\transpose$) on nearby regions (see \autocite{wand1994multivariate} for a more extensive explanation of the bandwidth matrix). 
There are many different ways of estimating the bandwidth matrix, ranging from simple reference rules like, e.g., Silverman's rule of thumb \autocite{silverman1986density} to more elaborate methods; see \autocite{turlach1993bandwidthselection, chiu1996comparative, jones1996brief, bashtannyk2001bandwidth, zambom2013review} for reviews of different bandwidth selection methods.

\cstart To estimate $\probabilitycond{\collision}{\situationinitial}$ of \cref{eq:probability collision expectation}, we need to draw samples from $\densityestcond{\situationfuture}{\situationinitial}$. \cend
Drawing samples from the estimated \ac{pdf} in \cref{eq:kde estimate} is straightforward: two random numbers are drawn, one to choose a random generator kernel out of the $\situationnumberof$ kernels that are used to construct the \ac{kde}, and one random number from that kernel.
Sampling from $\densityestcond{\situationfuture}{\situationinitial}$ works similarly, but instead of using an equal probability for each random generator kernel to be selected, different probabilities are used based on $\situationinitial$.
For more information on sampling from a conditional \ac{pdf} obtained using \ac{kde}, see \autocite{holmes2012fast, degelder2021conditional}.

\subsubsection{Assuming independence}
\label{sec:no special case}

Due to the curse of dimensionality \autocite{scott2015multivariate}, estimating $\density{\situationinitial, \situationfuture}$ with one \ac{kde} according to \cref{eq:kde estimate} becomes inaccurate if $\situationinitialdim + \situationfuturedim$ becomes large.
One option to avoid this curse of dimensionality is to assume that one or more parameters are independent of the other parameters. 
E.g., suppose that $\situationfuture\transpose=\begin{bmatrix}\situationfutureparta\transpose & \situationfuturepartb\transpose\end{bmatrix}$, such that $\situationfuturepartb$ is independent of $\situationinitial$ and $\situationfutureparta$.
Then we can write
\begin{equation}
	\density{\situationinitial, \situationfuture}
	= \density{\situationinitial, \situationfutureparta, \situationfuturepartb}
	= \density{\situationinitial, \situationfutureparta} \density{\situationfuturepartb}.
\end{equation}
In this case, we would need to estimate $\density{\situationinitial, \situationfutureparta}$ and $\density{\situationfuturepartb}$, which can be done in a similar manner as presented in \cref{sec:one kde}.
Because these two \acp{pdf} have fewer variables than $\density{\situationinitial, \situationfuture}$, the two estimated \acp{pdf} will suffer less from the curse of dimensionality \autocite{scott2015multivariate}.

Another option is to model $\densitycond{\situationfuture}{\situationinitial}$ as a cascade of conditional probabilities. 
For example, using the partitioning $\situationfuture\transpose=\begin{bmatrix}\situationfutureparta\transpose & \situationfuturepartb\transpose\end{bmatrix}$, $\densitycond{\situationinitial}{\situationfuture}$ can be approximated using two conditional densities:
\begin{equation}
	\densitycond{\situationfuture}{\situationinitial}
	= \densitycond{\situationfutureparta, \situationfuturepartb}{\situationinitial}
	= \densitycond{\situationfutureparta}{\situationfuturepartb, \situationinitial} \densitycond{\situationfuturepartb}{\situationinitial}
	\approx \densitycond{\situationfutureparta}{\situationfuturepartb} \densitycond{\situationfuturepartb}{\situationinitial}.
\end{equation}
This approximation is valid if $\situationfutureparta$ and $\situationinitial$ are \emph{conditionally independent given $\situationfuturepartb$} \autocite{nagler2016evading}.
The same partitioning can be applied to $\densitycond{\situationfutureparta}{\situationfuturepartb}$ and $\densitycond{\situationfuturepartb}{\situationinitial}$ until only two-dimensional \acp{pdf} need to be estimated.
Although this will lead to larger approximation errors, the lower-dimensional \acp{pdf} can be estimated more accurately. 
For more information on this approach, we refer the reader to \autocite{aas2009paircopula, nagler2016evading}.
\cstart Note that when relying on the assumption of independence or the assumption of conditional independence, these assumptions should be justified, e.g., through the use of some statistical tests like the Pearson's chi-squared test. \cend

\subsubsection{Reduce number of parameters using singular value decomposition}
\label{sec:parameter reduction}

Another way to avoid the curse of dimensionality is to use \iac{svd} \autocite{golub2013matrix} to reduce the number of parameters.
\cstart In the field of machine learning, \ac{pca} is commonly used for dimensionality reduction \autocite{abdi2010principal,hasan2021review} and \ac{pca} uses the \ac{svd}. \cend
With \iac{svd}, the parameters $\situationinitial$ and $\situationfuture$ are transformed into a lower-dimensional vector of parameters in such a way that the reduced vector of parameters describes as much of the variation as possible.
To do this, \iac{svd} is made of the matrix that contains all $\situationnumberof$ observed situations:
\begin{equation}
	\begin{bmatrix}
		\situationinitialinstance{1}-\situationinitialmean & \cdots & \situationinitialinstance{\situationnumberof}-\situationinitialmean \\
		\situationfutureinstance{1}-\situationfuturemean & \cdots & \situationfutureinstance{\situationnumberof}-\situationfuturemean
	\end{bmatrix} = \svdu \svds \svdv\transpose.
\end{equation}
Here, $\situationinitialmean=\frac{1}{\situationnumberof}\sum_{\situationindex=1}^{\situationnumberof}\situationinitialinstance{\situationindex}$ and $\situationfuturemean=\frac{1}{\situationnumberof}\sum_{\situationindex=1}^{\situationnumberof}\situationfutureinstance{\situationindex}$.
The matrices $\svdu \in \realnumbers^{\left(\situationinitialdim+\situationfuturedim\right)\times\left(\situationinitialdim+\situationfuturedim\right)}$ and $\svdv \in \realnumbers^{\situationnumberof \times \situationnumberof}$ are orthonormal, i.e., $\svdu^{-1}=\svdu\transpose$ and $\svdv^{-1}=\svdv\transpose$.
Moreover, $\svds\in\realnumbers^{\left(\situationinitialdim+\situationfuturedim\right)\times\situationnumberof}$ has only zeros except at the diagonal: the $(\svdindex,\svdindex)$-th element is $\svdsv{\svdindex}$, $\svdindex\in\{1,\ldots,\svdrank\}$ with  $\svdrank=\min(\situationinitialdim+\situationfuturedim, \situationnumberof)$, such that
\begin{equation}
	\svdsv{1} \geq \svdsv{2} \geq \ldots \geq \svdsv{\svdrank} \geq 0.
\end{equation}
Because these so-called singular values are in decreasing order, we can approximate $\situationinitial$ and $\situationfuture$ by setting $\svdsv{\svdindex}=0$ for $\svdindex > \dimension$ with $\dimension$ chosen\footnote{ We have $\dimension < \situationinitialdim+\situationfuturedim$, such that the dimension is reduced (from $\situationinitialdim+\situationfuturedim$ to $\dimension$) and we have $\dimension>\situationinitialdim$, such that the number of linear constraints in \cref{eq:linear constraint} ($\situationinitialdim$) is smaller than the number of variables ($\dimension$).} such that $\situationinitialdim < \dimension < \situationinitialdim+\situationfuturedim$:
\begin{equation}
	\label{eq:svd approximation}
	\begin{bmatrix}
		\situationinitialinstance{\situationindex} - \situationinitialmean \\
		\situationfutureinstance{\situationindex} - \situationfuturemean
	\end{bmatrix}
	= \sum_{\svdindex=1}^{\svdrank} \svdsv{\svdindex} \svdventry{\situationindex}{\svdindex} \svduvec{\svdindex}
	\approx \sum_{\svdindex=1}^{\dimension} \svdsv{\svdindex} \svdventry{\situationindex}{\svdindex} \svduvec{\svdindex},
	= \begin{bmatrix} \svduupperleft \\ \svdulowerleft \end{bmatrix} \svdsupperleft \svdvvecd{\situationindex},
\end{equation}
where $\svdventry{\situationindex}{\svdindex}$ is the $(\situationindex,\svdindex)$-th element of $\svdv$ and $\svduvec{\svdindex}$ is the $\svdindex$-th column of $\svdu$.
Moreover, $\svduupperleft$ is the $\situationinitialdim$-by-$\dimension$ upper left submatrix of $\svdu$, $\svdulowerleft$ is the $\situationfuturedim$-by-$\dimension$ lower left submatrix $\svdu$, $\svdsupperleft\in\realnumbers^{\dimension\times\dimension}$ is the diagonal matrix with the first $\dimension$ singular values on its diagonal, and $\svdvvecd{\situationindex}\transpose = \begin{bmatrix} \svdventry{\situationindex}{1} & \cdots & \svdventry{\situationindex}{\dimension} \end{bmatrix}$.
Thus, with $\situationinitialmean$, $\situationfuturemean$, $\svduupperleft$, $\svdulowerleft$, and $\svdsupperleft$, the $(\situationinitialdim+\situationfuturedim)$-dimensional vector $\begin{bmatrix}\situationinitialinstance{\situationindex}\transpose & \situationfutureinstance{\situationindex}\transpose\end{bmatrix}\transpose$ is approximated using the $\dimension$-dimensional vector $\svdvvecd{\situationindex}$.

Instead of estimating the \ac{pdf} of $\begin{bmatrix}\situationinitialinstance{\situationindex}\transpose & \situationfutureinstance{\situationindex}\transpose\end{bmatrix}\transpose$, we now estimate the \ac{pdf} of $\svdvvecd{\situationindex}$ using \ac{kde} as described in \cref{sec:one kde}.
\cstart Note that the choice of $\dimension$ includes a trade-off. 
Choosing $\dimension$ too small results in too much loss of detail, while choosing $\dimension$ too large will give accuracy problems when estimating the \ac{pdf} of the new parameters.
For more information on choosing an appropriate value of $\dimension$, we refer the reader to \autocite{degelder2021generation}. \cend
To sample from $\densityestcond{\situationfuture}{\situationinitial}$, we can sample from the estimated distribution of $\svdvvecd{\situationindex}$.
Because \cref{eq:svd approximation} is a linear mapping, the sample $\svdvvecsymbol$ that is drawn from the estimated distribution of $\svdvvecd{\situationindex}$ is subject to a linear constraint:
\begin{equation}
	\label{eq:linear constraint}
	\svduupperleft \svdsupperleft \svdvvecsymbol = \situationinitial - \situationinitialmean.
\end{equation}
In \autocite{degelder2021conditional}, an algorithm is provided for sampling from \iac{kde} with a Gaussian kernel of \cref{eq:kernel initial future} such that the resulting samples are subject to a linear constraint such as \cref{eq:linear constraint}.

\subsection{Estimate $\probabilitycond{\collision}{\situationinitial}$ using a Monte Carlo simulation}
\label{sec:estimate collision}

Monte Carlo simulations are used to estimate $\probabilitycond{\collision}{\situationinitial}$, i.e., the probability of an event $\collision$ given the initial situation described by $\situationinitial$.
The details of the simulation depend on the actual application. 
For example, if the goal of our \ac{ssm} is to evaluate the risk that a human-driven vehicle collides, the simulation should involve a human driving behavior model. 
On the other hand, if the goal is to evaluate the risk of a crash when \iac{ads} is controlling the vehicle, the simulation should include the model of this \ac{ads}.

A straightforward way to compute $\probabilitycond{\collision}{\situationinitial}$ is to repeat a certain number of simulations with the same $\situationinitial$ and count the number of simulations that result in the event $\collision$.
If $\numberofsimulations$ denotes the number of simulations and $\numberofcollisions$ is the number of events $\collision$, then $\probabilitycond{\collision}{\situationinitial}$ could be estimated using
\begin{equation}
	\label{eq:binomial estimation}
	\probabilityestcond{\collision}{\situationinitial}
	= \frac{\numberofcollisions}{\numberofsimulations}.
\end{equation}

An important choice for estimating $\probabilitycond{\collision}{\situationinitial}$ is the number of simulations, $\numberofsimulations$.
One approach is to keep increasing $\numberofsimulations$ until there is enough confidence in the estimation of \cref{eq:binomial estimation}.
For example, the Clopper-Pearson interval \autocite{clopper1934use} or the Wilson score interval \autocite{wilson1927probable} can be used to determine the confidence of the estimation of \cref{eq:binomial estimation}.
A disadvantage of \cref{eq:binomial estimation} is that only the fact whether the event $\collision$ occurred or not is used, while the simulation provides more information, such as the minimum distance between two objects or the impact speed in case of a crash.
Therefore, we provide an alternative approach to estimate $\probabilitycond{\collision}{\situationinitial}$.

For the alternative approach, let us assume that one simulation run provides more information than just the fact that the event $\collision$ occurred or not.
Let $\simulationresult \in \realnumbers^{\dimsimulationresult}$ be a continuous variable representing the result of a simulation run and let $\spacecollision$ denote the set of possible simulation results in which the event $\collision$ occurred. 
Thus, $\simulationresult \in \spacecollision$ if and only if the simulation results in the event $\collision$.
We assume $\spacecollision$ is known; see, e.g., the example in \cref{sec:wang stamatiadis replicate}.
Therefore, we have
\begin{equation}
	\probabilitycond{\collision}{\situationinitial}
	= \probabilitycond{\simulationresult \in \spacecollision}{\situationinitial}
	= \int_{\spacecollision} \densitycond{\simulationresult}{\situationinitial} \ud \simulationresult.
\end{equation}
Similar as with the estimation of $\density{\situationinitial, \situationfuture}$ in \cref{sec:estimate future}, we employ \ac{kde} to estimate $\densitycond{\simulationresult}{\situationinitial}$:
\begin{equation}
	\label{eq:kde simulation result}
	\densityestcond{\simulationresult}{\situationinitial}
	= \frac{1}{\numberofsimulations} 
	\sum_{\simulationindex=1}^{\numberofsimulations} \kernelfuncnormalized{\simulationbandwidth}{\simulationinstance{\simulationindex} - \simulationresult},
\end{equation}
where $\simulationinstance{\simulationindex}$ denotes the result of the $\simulationindex$-th simulation and $\simulationbandwidth$ denotes an appropriate bandwidth matrix.
The kernel function $\kernelfuncnormalized{\simulationbandwidth}{\cdot}$ is similarly defined as \cref{eq:kernel initial future}.
We can now estimate $\probabilitycond{\collision}{\situationinitial}$ by substituting $\densityestcond{\simulationresult}{\situationinitial}$ of \cref{eq:kde simulation result} for $\densitycond{\simulationresult}{\situationinitial}$:
\begin{equation}
	\label{eq:estimate probability of collision}
	\probabilityestcond{\collision}{\situationinitial}
	= \probabilityestcond{\simulationresult \in \spacecollision}{\situationinitial}
	= \int_{\spacecollision} \densityestcond{\simulationresult}{\situationinitial} \ud \simulationresult
	=\frac{1}{\numberofsimulations}
	\sum_{\simulationindex=1}^{\numberofsimulations} \int_{\spacecollision}
	\kernelfuncnormalized{\simulationbandwidth}{\simulationinstance{\simulationindex} - \simulationresult} \ud \simulationresult.
\end{equation}

Similar as with \cref{eq:binomial estimation}, we need to choose the number of simulations $\numberofsimulations$.
Our proposal is to keep increasing $\numberofsimulations$ until \cstart the uncertainty of the estimated probability $\probabilityestcond{\simulationresult \in \spacecollision}{\situationinitial}$ is below a certain threshold. 
As a measure for the uncertainty of the estimated probability $\probabilityestcond{\simulationresult \in \spacecollision}{\situationinitial}$, we use the variance of $\probabilityestcond{\simulationresult \in \spacecollision}{\situationinitial}$. 
Thus, we keep increasing $\numberofsimulations$ until \cend the variance of $\probabilityestcond{\simulationresult \in \spacecollision}{\situationinitial}$ is below a threshold $\simulationthreshold>0$.
The variance follows from \autocite{nadaraya1964some}:
\begin{equation}
	\label{eq:variance estimation}
	\variance{\probabilityestcond{\simulationresult \in \spacecollision}{\situationinitial}}
	= \frac{\probabilitycond{\simulationresult \in \spacecollision}{\situationinitial}
		\left( 1-\probabilitycond{\simulationresult \in \spacecollision}{\situationinitial} \right)}{\numberofsimulations}.
\end{equation}
Because $\probabilitycond{\simulationresult \in \spacecollision}{\situationinitial}$ is unknown, \cstart we cannot directly use \cref{eq:variance estimation}.
Instead, \cend we \cstart substitute \cend the estimated counterpart of \cref{eq:estimate probability of collision} \cstart for $\probabilitycond{\simulationresult \in \spacecollision}{\situationinitial}$\cend.
Thus, $\numberofsimulations$ is increased until the following condition is met:
\begin{equation}
	\label{eq:condition stop simulations}
	\frac{\probabilityestcond{\simulationresult \in \spacecollision}{\situationinitial}
		\left( 1-\probabilityestcond{\simulationresult \in \spacecollision}{\situationinitial} \right)}{\numberofsimulations}
	< \simulationthreshold.
\end{equation}

\subsection{Regression for real-time estimation of $\probabilitycond{\collision}{\situationinitial}$}
\label{sec:final metric calculation}

To evaluate the risk measure during real-time operation of the ego vehicle, the expression of \cref{eq:estimate probability of collision} is problematic, because it would require $\numberofsimulations$ simulations.
Even if the calculation is accelerated using a technique such as importance sampling, it might take too long.
Therefore, we propose to evaluate \cref{eq:estimate probability of collision} only for some fixed $\left\{\situationinitialinstance{\situationindexdesign}'\right\}_{\situationindexdesign=1}^{\numberofdesignpoints}$.
Next, regression is used to estimate \cref{eq:estimate probability of collision}.
One option is to choose a parametric model, e.g., a logistic model, and estimate the parameters of the model using $\left\{\left(\situationinitialinstance{\situationindexdesign}',\probabilityestcond{\collision}{\situationinitialinstance{\situationindexdesign}'}\right)\right\}_{\situationindexdesign=1}^{\numberofdesignpoints}$.
Up to our knowledge, however, there is no good reason to assume a particular parametric model, so we use a non-parametric regression technique to estimate \cref{eq:estimate probability of collision}.
More specifically, we use the \acf{nw} kernel estimator \autocite{wasserman2006nonparametric}, because it automatically smooths the data (as is demonstrated in \cref{sec:wang stamatiadis comparison}) and the approximation is guaranteed to give a number between 0 and 1, also when extrapolating the data.
The \ac{nw} kernel estimator is given by: 
\begin{equation}
	\label{eq:nadaraya watson}
	\probabilityestcond{\collision}{\situationinitial}
	\approx \frac{ \sum_{\situationindexdesign=1}^{\numberofdesignpoints}
		\kernelfuncnormalized{\bandwidthnw}{\situationinitial - \situationinitialinstance{\situationindexdesign}'}
		\probabilityestcond{\collision}{\situationinitialinstance{\situationindexdesign}'}
	}{\sum_{\situationindexdesign=1}^{\numberofdesignpoints}
		\kernelfuncnormalized{\bandwidthnw}{\situationinitial - \situationinitialinstance{\situationindexdesign}'}}.
\end{equation}
Here, $\probabilityestcond{\collision}{\situationinitialinstance{\situationindexdesign}'}$ is based on \cref{eq:estimate probability of collision} and $\kernelfuncnormalized{\bandwidthnw}{\cdot}$ represents the Gaussian kernel given by \cref{eq:kernel initial future}.
Two important choices have to be made: The choice of $\left\{\situationinitialinstance{\situationindexdesign}'\right\}_{\situationindexdesign=1}^{\numberofdesignpoints}$ for which to evaluate \cref{eq:estimate probability of collision} and the choice of the bandwidth matrix $\bandwidthnw$.
We suggest to base the design points $\left\{\situationinitialinstance{\situationindexdesign}'\right\}_{\situationindexdesign=1}^{\numberofdesignpoints}$ on the data that is used to estimate $\densitycond{\situationfuture}{\situationinitial}$ in \cref{sec:estimate future}, i.e., $\left\{\situationinitialinstance{\situationindex}\right\}_{\situationindex=1}^{\situationnumberof}$, such that all $\situationinitialinstance{\situationindex}$ have at least one design point $\situationinitialinstance{\situationindexdesign}'$ nearby.
In other words, $\left\{\situationinitialinstance{\situationindexdesign}'\right\}_{\situationindexdesign=1}^{\numberofdesignpoints}$ is chosen such that
\begin{equation}
	\label{eq:design points distance}
	\min_{\situationindexdesign} 
	\left( \situationinitialinstance{\situationindex} - \situationinitialinstance{\situationindexdesign}' \right)\transpose
	\weightmatrix 
	\left( \situationinitialinstance{\situationindex} - \situationinitialinstance{\situationindexdesign}' \right)
	\leq 1,
	\quad \forall \situationindex \in \{1, \ldots, \situationnumberof\},
\end{equation}
where $\weightmatrix$ denotes a weighting matrix. 
Note that if $\weightmatrix$ is the identity matrix, then \cref{eq:design points distance} calculates the minimum squared Euclidean distance.
In general, $\weightmatrix$ is a diagonal matrix.
Choosing the diagonal elements of $\weightmatrix$ is a trade-off; if the elements are too large, then too many details are lost in the approximation of \cref{eq:nadaraya watson}; if the elements are too small, it takes too long to evaluate \cref{eq:estimate probability of collision} $\numberofdesignpoints$ times, as $\numberofdesignpoints$ increases for lower diagonal elements of $\weightmatrix$.
The bandwidth matrix $\bandwidthnw$ might be based on $\weightmatrix$, e.g., $\bandwidthnw=\weightmatrix^{-1}$.
Alternatively, $\bandwidthnw$ might be based on the measurement uncertainty of $\situationinitial$ if this measurement uncertainty is significant, where a larger $\bandwidthnw$ applies in case of a larger measurement uncertainty of $\situationinitial$.
Note that if $\bandwidthnw$ is a diagonal matrix with positive values on the diagonal that are close to zero, then the \ac{nw} kernel estimation of \cref{eq:nadaraya watson} acts like nearest-neighbor interpolation.

\section{Case study}
\label{sec:case study}

In the first part of the case study, we illustrate that the \ac{ourmethod} method generalizes the \ac{ssm} proposed by \textcite{wang2014evaluation}.
Here, we also demonstrate the effect of $\simulationthreshold$ on the accuracy of the \ac{ssm} derived by the \ac{ourmethod} method and we show the difference between $\probabilityestcond{\collision}{\situationinitial}$ of \cref{eq:estimate probability of collision} and the approximation of $\probabilityestcond{\collision}{\situationinitial}$ using the \ac{nw} kernel estimator of \cref{eq:nadaraya watson}.
In \cref{sec:ngsim metric}, we demonstrate how the \ac{ourmethod} method can be used to create a new \ac{ssm} that calculates the risk of a crash in a longitudinal interaction between two vehicles.
The \ac{ssm} derived in \cref{sec:ngsim metric} is qualitatively analyzed in \cref{sec:analyzing ngsim metric}.
To also quantitatively analyze \acp{ssm}, \textcite{mullakkal2017comparative} proposed a benchmarking method which we apply in \cref{sec:trends}.

\subsection{Comparison with Wang and Stamatiadis' measure}
\label{sec:wang stamatiadis}

\textcite{wang2014evaluation} provide \iac{ssm}, which we denote by $\wangstamatiadis$, that calculates the probability of a crash under certain assumptions. 
We first explain how $\wangstamatiadis$ is calculated. 
Next, \cref{sec:wang stamatiadis replicate} shows how to estimate this \ac{ssm} using our method.
In \cref{sec:wang stamatiadis comparison}, we illustrate the results of both.

\subsubsection{Measure of Wang and Stamatiadis}
\label{sec:wang stamatiadis explanation}

The \ac{ssm} $\wangstamatiadis$ calculates the probability of a crash of the ego vehicle and the leading vehicle, where the ego vehicle is following an initially slower driving leading vehicle.
The \ac{ssm} $\wangstamatiadis$ is based on the following assumptions \autocite{wang2014evaluation}:
\begin{itemize}
	\item the leading vehicle keeps a constant speed;
	\item the (driver of the) ego vehicle starts to brake after its reaction time, denoted by $\timereact$;
	\item based on \autocite{green2000long}, the reaction time $\timereact$ is distributed according to a log-normal distribution, such that the mean is \SI{0.92}{\second} and the standard deviation is \SI{0.28}{\second};
	\item when the ego vehicle reacts, it brakes with its \ac{madr}, denoted by $\accelerationmax$; and
	\item $\accelerationmax$ is distributed according to a truncated normal distribution with a mean of $\SI{9.7}{\meter\per\second\squared}$, a standard deviation of $\SI{1.3}{\meter\per\second\squared}$, a lower bound of $\lowerbound=\SI{4.2}{\meter\per\second\squared}$ \autocite{cunto2008assessing}, and an upper bound of $\upperbound=\SI{12.7}{\meter\per\second\squared}$ \autocite{cunto2008assessing}.
\end{itemize}
To calculate $\wangstamatiadis$ at a given time $\time$, the speed difference between the ego vehicle and the leading vehicle, $\speeddifference{\time}$, and the \ac{ttc}, $\ttc{\time}$, are used.
Note that $\ttc{\time}$ is the ratio of the gap, $\gap{\time}$, between the ego vehicle and the leading vehicle and $\speeddifference{\time}$.
If $\speeddifference{\time} \leq 0$, then the ego vehicle drives slower and there is no risk of a future crash according to \textcite{wang2014evaluation}, so $\wangstamatiadis(\time)=0$.
Given $\accelerationmax$, the driver of the ego vehicle needs to react within
\begin{equation}
	\timemaxreact(\time) = \ttc{\time} - \frac{\speeddifference{\time}}{2 \accelerationmax}
\end{equation}
in order to avoid a crash. 
Using the distributions of $\accelerationmax$ and $\timereact$, we can calculate the probability that this is the case, resulting in:
\begin{equation}
	\label{eq:ws}
	\wangstamatiadis(\time) = \begin{cases}
		0 & \text{if}\quad \speeddifference{\time} \leq 0 \\
		\int_{\lowerboundadapted}^{\upperbound}
		\int_{0}^{\timemaxreact(\time)}
		\density{\timereact} \density{\accelerationmax} \ud \timereact \ud \accelerationmax
		& \text{if}\quad \speeddifference{\time} > 0 \wedge \frac{\speeddifference{\time}}{2\ttc{\time}} < \upperbound \\
		1 & \text{otherwise}
	\end{cases},
\end{equation}
with $\lowerboundadapted=\max \left( \lowerbound, \frac{\speeddifference{\time}}{2\ttc{\time}}\right)$,
$\density{\timereact}$ is the log-normal probability density of $\timereact$, and $\density{\accelerationmax}$ is the truncated normal probability density of $\accelerationmax$.

\subsubsection{Replicating Wang and Stamatiadis' measure}
\label{sec:wang stamatiadis replicate}

Because $\wangstamatiadis$ is based on $\speeddifference{\time}$ and $\ttc{\time}$, these two variables are also used by the \ac{ourmethod} method to describe the initial situation:
\begin{equation}
	\label{eq:situation initial ws}
	\situationinitial\transpose(\time) = \begin{bmatrix}
		\speeddifference{\time} & \ttc{\time}
	\end{bmatrix}.
\end{equation}
The leading vehicle is assumed to have a constant speed, so $\situationinitial(\time)$ of \cref{eq:situation initial ws} already describes the future situation of the leading vehicle.
Therefore, there is no need to estimate $\densitycond{\situationfuture}{\situationinitial}$.
At the start of each simulation run, the driver of the ego vehicle is not braking. 
After the reaction time $\timereact$, the driver starts braking with $\accelerationmax$.
The random parameters $\timereact$ and $\accelerationmax$ are similarly distributed as described in \cref{sec:wang stamatiadis explanation}.

Since we are interested in the probability of a crash, the event $\collision$ denotes a crash.
A simulation run ends if either the ego vehicle and the leading vehicle are colliding or if the gap between the ego vehicle and the leading vehicle is not decreasing. 
Depending on the reason for a simulation run to end, we consider the following result:
\begin{itemize}
	\item If the ego vehicle and the leading vehicle are colliding, we are interested in the ``severity'' of the crash. 
	This is expressed using the speed difference: $\speedlead{\timeend} - \speedego{\timeend}$, where $\timeend$ denotes the final time of the simulation run.
	\item If there is no crash, we are interested in how close the two vehicles came.
	Therefore, the minimum gap is used, which is $\gap{\timeend}$.
\end{itemize}
Thus, we have:
\begin{equation}
	\label{eq:simulation result}
	\simulationresult = \begin{cases}
		\speedlead{\timeend} - \speedego{\timeend} & \text{if crash} \\
		\gap{\timeend} & \text{otherwise}
	\end{cases}.
\end{equation}
Clearly, $\simulationresult\leq0$ indicates a crash, so $\spacecollision=(-\infty, 0]$.
The minimum number of simulations to estimate $\probabilitycond{\collision}{\situationinitial}$ is set to 10. 
The number of simulations is further increased until the condition in \cref{eq:condition stop simulations} with $\simulationthreshold=0.2$ or $\simulationthreshold=0.02$ is met.
For the design points  $\left\{\situationinitialinstance{\situationindexdesign}'\right\}_{\situationindexdesign=1}^{\numberofdesignpoints}$, we use a rectangular grid with $\speeddifferencesymbol$ ranging from \SI{0}{\meter\per\second} till \SI{40}{\meter\per\second} with steps of \SI{2}{\meter\per\second} and $\ttcsymbol$ ranging from \SI{0.5}{\second} till \SI{4}{\second} in steps of \SI{0.1}{\second}.
Thus, $\numberofdesignpoints=21 \cdot 36=756$.
For $\bandwidthnw$, a diagonal matrix is chosen with the diagonal elements corresponding to the square of the step size of the grid, i.e., \SI{4}{\meter\squared\per\second\squared} and \SI{0.01}{\second\squared}.

\subsubsection{Comparison}
\label{sec:wang stamatiadis comparison}

\Cref{fig:ws comparison} shows the results of the comparison between the measure of \textcite{wang2014evaluation} and the measure derived using the \ac{ourmethod} method described in \cref{sec:method}.
The black lines in \cref{fig:ws comparison} denote $\wangstamatiadis$ of \cref{eq:ws}.
These lines show that for lower values of $\ttcsymbol$, $\wangstamatiadis$ increases.
Also, for increasing values of $\speeddifferencesymbol$ (solid, dashed, and dotted lines), the risk measure $\wangstamatiadis$ increases.
Both these observations match the intuition that a lower \ac{ttc} and a higher speed difference are less safe.

The gray lines in \cref{fig:ws comparison} denote $\probabilityestcond{\collision}{\situationinitial}$.
In \cref{fig:ws comparison coarse,fig:ws comparison fine}, $\probabilityestcond{\collision}{\situationinitial}$ of \cref{eq:estimate probability of collision} is used and the gray lines in \cref{fig:ws comparison coarse nw,fig:ws comparison fine nw} represent the approximation of $\probabilityestcond{\collision}{\situationinitial}$ using the \ac{nw} kernel estimator of \cref{eq:nadaraya watson}.
\Cref{fig:ws comparison} illustrates that $\probabilityestcond{\collision}{\situationinitial}$ follows the same trend as $\wangstamatiadis$.
\Cref{fig:ws comparison} also illustrates the effect of the choice of the threshold $\simulationthreshold$.
In general, for a lower value of $\simulationthreshold$, the number of simulations $\numberofsimulations$ used in \cref{eq:kde simulation result} is higher. 
As a result, it can be expected that the estimation $\probabilityestcond{\collision}{\situationinitial}$ is closer to $\probabilitycond{\collision}{\situationinitial}$ (cf.\ \cref{eq:variance estimation}).
A comparison of \cref{fig:ws comparison coarse} ($\simulationthreshold=0.2$) and \cref{fig:ws comparison fine} ($\simulationthreshold=0.02$) demonstrates this effect.
\Cref{fig:ws comparison} further illustrates the regression using the \ac{nw} kernel estimator: the gray lines in \cref{fig:ws comparison coarse nw,fig:ws comparison fine nw} can be seen as smoothed versions of the gray lines in \cref{fig:ws comparison coarse,fig:ws comparison fine}, respectively.

\setlength{\figurewidth}{.47\linewidth}
\setlength{\figureheight}{.7\figurewidth}
\begin{figure}[t]
	\centering
	\begin{subfigure}{.49\linewidth}
		\centering
\begin{tikzpicture}

\definecolor{darkgray176}{RGB}{176,176,176}

\begin{axis}[
height=\figureheight,
scaled y ticks=false,
tick align=outside,
tick pos=left,
width=\figurewidth,
x grid style={darkgray176},
xlabel={$\ttcsymbol$ [\si{\second}]},
xmajorgrids,
xmin=0.5, xmax=4,
xtick style={color=black},
xticklabel style={align=center},
y grid style={darkgray176},
ylabel={Probability of a crash},
ymajorgrids,
ymin=0, ymax=1,
ytick style={color=black},
yticklabel style={/pgf/number format/fixed,/pgf/number format/precision=3}
]
\addplot [very thick, black]
table {%
0.5 1
0.6 0.999999998869978
0.7 0.99999857290027
0.8 0.999897180595153
0.9 0.998283318978398
1 0.98817993034985
1.1 0.954499368567494
1.2 0.88238322312223
1.3 0.770109264089317
1.4 0.631738967436461
1.5 0.488426235243335
1.6 0.358220116360871
1.7 0.251152278540543
1.8 0.16961171159941
1.9 0.11109898777971
2 0.0710131882324527
2.1 0.0445256450697417
2.2 0.027506842417679
2.3 0.0168046929172315
2.4 0.0101836167889724
2.5 0.00613677052673833
2.6 0.00368492901741024
2.7 0.00220845719489648
2.8 0.00132282408579654
2.9 0.000792755323229866
3 0.000475750748254011
3.1 0.000286106476016124
3.2 0.00017251527884421
3.3 0.000104345369502878
3.4 6.33312409136222e-05
3.5 3.85817786129339e-05
3.6 2.35971162777515e-05
3.7 1.44916803196393e-05
3.8 8.93745841601401e-06
3.9 5.53582162532429e-06
4 3.44387841788585e-06
};
\addplot [very thick, gray]
table {%
0.5 1
0.6 1
0.7 1
0.8 1
0.9 0.998579279836947
1 1
1.1 0.999976558774051
1.2 0.868865749795263
1.3 0.852852356550611
1.4 0.717865020569852
1.5 0.721213798180494
1.6 0.478336687506885
1.7 0.444413179577576
1.8 7.88520220229572e-09
1.9 0.365719892486444
2 0.319415876979553
2.1 0.000608384122774847
2.2 0
2.3 0.000269177935865988
2.4 8.49805841807161e-09
2.5 2.51080295443629e-09
2.6 5.55111512312578e-18
2.7 0
2.8 0
2.9 0
3 6.20632229220153e-08
3.1 0
3.2 0
3.3 0
3.4 0
3.5 0
3.6 1.97786231836972e-14
3.7 2.57405208259343e-14
3.8 0
3.9 0
4 0
};
\addplot [very thick, black, dashed]
table {%
0.5 1
0.6 1
0.7 1
0.8 1
0.9 1
1 0.999999999348448
1.1 0.99999950235446
1.2 0.999973209728235
1.3 0.999598995569285
1.4 0.997158301080912
1.5 0.987738504067729
1.6 0.962905518012083
1.7 0.913879059405361
1.8 0.836569727613501
1.9 0.734465168099655
2 0.617381972771856
2.1 0.497529753222982
2.2 0.385599610401107
2.3 0.288575568937833
2.4 0.209429765332663
2.5 0.14799359805209
2.6 0.102200293054103
2.7 0.0691792751327623
2.8 0.0460033719080388
2.9 0.0300971283204075
3 0.0193911312382948
3.1 0.0123154370919198
3.2 0.00772044024779683
3.3 0.004785490293998
3.4 0.0029387118341998
3.5 0.00179149195992223
3.6 0.00108628571142166
3.7 0.000656315154883846
3.8 0.000395725169891836
3.9 0.000238431669817296
4 0.000143716534243055
};
\addplot [very thick, gray, dashed]
table {%
0.5 1
0.6 1
0.7 1
0.8 1
0.9 1
1 1
1.1 1
1.2 1
1.3 1
1.4 1
1.5 0.999999999999743
1.6 1
1.7 0.999994170494515
1.8 0.629814188544941
1.9 0.632840713530643
2 0.626925016445234
2.1 0.648412310932713
2.2 0.392635285415616
2.3 0.302606358898226
2.4 0.401482913116048
2.5 0.200210285651481
2.6 0.208429826073663
2.7 0.135915135881802
2.8 4.90188490780974e-06
2.9 3.41915520752911e-06
3 1.01069840885426e-08
3.1 1.50613018723433e-08
3.2 8.13403233657084e-12
3.3 1.7733384760632e-10
3.4 0.0121889042499522
3.5 0
3.6 3.1635165376187e-09
3.7 0
3.8 3.18467474613726e-14
3.9 0
4 0
};
\addplot [very thick, black, dotted]
table {%
0.5 1
0.6 1
0.7 1
0.8 1
0.9 1
1 1
1.1 1
1.2 1
1.3 0.999999999999999
1.4 0.999999999377245
1.5 0.999999657092481
1.6 0.999985156187676
1.7 0.999809817862939
1.8 0.998773565728781
1.9 0.994896330962281
2 0.984383486717914
2.1 0.962047971359528
2.2 0.922923942475201
2.3 0.864322979544098
2.4 0.787151363442989
2.5 0.69581547694205
2.6 0.596926356773646
2.7 0.497553920246217
2.8 0.403757956805445
2.9 0.319768708556622
3 0.247822552515213
3.1 0.18844874034795
3.2 0.140960526053158
3.3 0.10396026119255
3.4 0.0757525703567505
3.5 0.0546306025901766
3.6 0.0390439069936283
3.7 0.0276755214076069
3.8 0.0194589062185967
3.9 0.0135605539107035
4 0.00934782588289695
};
\addplot [very thick, gray, dotted]
table {%
0.5 1
0.6 1
0.7 1
0.8 1
0.9 1
1 1
1.1 1
1.2 1
1.3 1
1.4 1
1.5 1
1.6 1
1.7 1
1.8 1
1.9 0.920603808897014
2 1
2.1 0.941526887743996
2.2 0.850685562956434
2.3 0.870055348570052
2.4 0.748100596981206
2.5 0.810731537139966
2.6 0.732482462112221
2.7 0.539323813755196
2.8 0.689748240130972
2.9 0.383160263046252
3 0.245469156128095
3.1 0.258514053604671
3.2 0.155760461493876
3.3 1.94283089616221e-11
3.4 0.220559736447763
3.5 9.6596313142211e-05
3.6 0.0980533107905212
3.7 0.00692347867034816
3.8 0
3.9 0.104669763926665
4 0
};
\end{axis}

\end{tikzpicture}
		\caption{Comparison of $\wangstamatiadis$ of \cref{eq:ws} (black) and $\probabilityestcond{\collision}{\situationinitial}$ of \cref{eq:estimate probability of collision} (gray) with $\simulationthreshold=0.2$.}
		\label{fig:ws comparison coarse}
	\end{subfigure}\hfill
	\begin{subfigure}{.49\linewidth}
		\centering
\begin{tikzpicture}

\definecolor{darkgray176}{RGB}{176,176,176}

\begin{axis}[
height=\figureheight,
scaled y ticks=false,
tick align=outside,
tick pos=left,
width=\figurewidth,
x grid style={darkgray176},
xlabel={$\ttcsymbol$ [\si{\second}]},
xmajorgrids,
xmin=0.5, xmax=4,
xtick style={color=black},
xticklabel style={align=center},
y grid style={darkgray176},
ylabel={Probability of a crash},
ymajorgrids,
ymin=0, ymax=1,
ytick style={color=black},
yticklabel style={/pgf/number format/fixed,/pgf/number format/precision=3}
]
\addplot [very thick, black]
table {%
0.5 1
0.6 0.999999998869978
0.7 0.99999857290027
0.8 0.999897180595153
0.9 0.998283318978398
1 0.98817993034985
1.1 0.954499368567494
1.2 0.88238322312223
1.3 0.770109264089317
1.4 0.631738967436461
1.5 0.488426235243335
1.6 0.358220116360871
1.7 0.251152278540543
1.8 0.16961171159941
1.9 0.11109898777971
2 0.0710131882324527
2.1 0.0445256450697417
2.2 0.027506842417679
2.3 0.0168046929172315
2.4 0.0101836167889724
2.5 0.00613677052673833
2.6 0.00368492901741024
2.7 0.00220845719489648
2.8 0.00132282408579654
2.9 0.000792755323229866
3 0.000475750748254011
3.1 0.000286106476016124
3.2 0.00017251527884421
3.3 0.000104345369502878
3.4 6.33312409136222e-05
3.5 3.85817786129339e-05
3.6 2.35971162777515e-05
3.7 1.44916803196393e-05
3.8 8.93745841601401e-06
3.9 5.53582162532429e-06
4 3.44387841788585e-06
};
\addplot [very thick, gray]
table {%
0.5 1
0.6 1
0.7 1
0.8 1
0.9 1
1 0.999999999997186
1.1 0.999999999998239
1.2 0.895411402324518
1.3 0.786803031476875
1.4 0.666042217005033
1.5 0.469533719945522
1.6 0.441686255743984
1.7 0.301183855962726
1.8 0.261377436262721
1.9 0.103752520425265
2 0.0717860495716075
2.1 1.96509475358653e-15
2.2 8.27609730746914e-10
2.3 0.102183624532358
2.4 2.49272558368596e-06
2.5 2.83644829845997e-11
2.6 0
2.7 0.000814807379380539
2.8 0.00339320800082153
2.9 2.72403338913429e-07
3 0
3.1 0
3.2 0
3.3 0
3.4 4.10782519111308e-16
3.5 0
3.6 6.32827124036339e-16
3.7 0
3.8 0
3.9 0
4 0
};
\addplot [very thick, black, dashed]
table {%
0.5 1
0.6 1
0.7 1
0.8 1
0.9 1
1 0.999999999348448
1.1 0.99999950235446
1.2 0.999973209728235
1.3 0.999598995569285
1.4 0.997158301080912
1.5 0.987738504067729
1.6 0.962905518012083
1.7 0.913879059405361
1.8 0.836569727613501
1.9 0.734465168099655
2 0.617381972771856
2.1 0.497529753222982
2.2 0.385599610401107
2.3 0.288575568937833
2.4 0.209429765332663
2.5 0.14799359805209
2.6 0.102200293054103
2.7 0.0691792751327623
2.8 0.0460033719080388
2.9 0.0300971283204075
3 0.0193911312382948
3.1 0.0123154370919198
3.2 0.00772044024779683
3.3 0.004785490293998
3.4 0.0029387118341998
3.5 0.00179149195992223
3.6 0.00108628571142166
3.7 0.000656315154883846
3.8 0.000395725169891836
3.9 0.000238431669817296
4 0.000143716534243055
};
\addplot [very thick, gray, dashed]
table {%
0.5 1
0.6 1
0.7 1
0.8 1
0.9 1
1 1
1.1 1
1.2 1
1.3 1
1.4 1
1.5 1
1.6 1
1.7 0.999789364091494
1.8 0.995597404480383
1.9 0.829864569965757
2 0.658238321475429
2.1 0.460751290373472
2.2 0.508308321463212
2.3 0.259072537384431
2.4 0.250765292476781
2.5 0.00666144869988307
2.6 0.13978852218325
2.7 0.0974798294702629
2.8 3.61044527608101e-14
2.9 0
3 5.83776535068026e-07
3.1 1.44995127016045e-14
3.2 8.99280649946377e-15
3.3 0
3.4 0
3.5 0
3.6 0.000262860323517888
3.7 0.0285714707708168
3.8 0
3.9 0
4 0
};
\addplot [very thick, black, dotted]
table {%
0.5 1
0.6 1
0.7 1
0.8 1
0.9 1
1 1
1.1 1
1.2 1
1.3 0.999999999999999
1.4 0.999999999377245
1.5 0.999999657092481
1.6 0.999985156187676
1.7 0.999809817862939
1.8 0.998773565728781
1.9 0.994896330962281
2 0.984383486717914
2.1 0.962047971359528
2.2 0.922923942475201
2.3 0.864322979544098
2.4 0.787151363442989
2.5 0.69581547694205
2.6 0.596926356773646
2.7 0.497553920246217
2.8 0.403757956805445
2.9 0.319768708556622
3 0.247822552515213
3.1 0.18844874034795
3.2 0.140960526053158
3.3 0.10396026119255
3.4 0.0757525703567505
3.5 0.0546306025901766
3.6 0.0390439069936283
3.7 0.0276755214076069
3.8 0.0194589062185967
3.9 0.0135605539107035
4 0.00934782588289695
};
\addplot [very thick, gray, dotted]
table {%
0.5 1
0.6 1
0.7 1
0.8 1
0.9 1
1 1
1.1 1
1.2 1
1.3 1
1.4 1
1.5 1
1.6 1
1.7 1
1.8 1
1.9 1
2 1
2.1 0.999999999622026
2.2 0.999950566082014
2.3 0.888021091989175
2.4 0.887411833772842
2.5 0.676890676505935
2.6 0.640539619951081
2.7 0.479465658859252
2.8 0.422356040807247
2.9 0.385931457927929
3 0.0015545552403848
3.1 0.203448914542212
3.2 0.197857761239706
3.3 0.109464326981113
3.4 0.0963830232361504
3.5 0.123867940392607
3.6 9.4460569619037e-07
3.7 0
3.8 1.57278251850856e-05
3.9 0.0285714285728185
4 0
};
\end{axis}

\end{tikzpicture}
		\caption{Comparison of $\wangstamatiadis$ of \cref{eq:ws} (black) and $\probabilityestcond{\collision}{\situationinitial}$ of \cref{eq:estimate probability of collision} (gray) with $\simulationthreshold=0.02$.}
		\label{fig:ws comparison fine}
	\end{subfigure}
	\begin{subfigure}{.49\linewidth}
		\centering
\begin{tikzpicture}

\definecolor{darkgray176}{RGB}{176,176,176}

\begin{axis}[
height=\figureheight,
scaled y ticks=false,
tick align=outside,
tick pos=left,
width=\figurewidth,
x grid style={darkgray176},
xlabel={$\ttcsymbol$ [\si{\second}]},
xmajorgrids,
xmin=0.5, xmax=4,
xtick style={color=black},
xticklabel style={align=center},
y grid style={darkgray176},
ylabel={Probability of collision},
ymajorgrids,
ymin=0, ymax=1,
ytick style={color=black},
yticklabel style={/pgf/number format/fixed,/pgf/number format/precision=3}
]
\addplot [very thick, black]
table {%
0.5 1
0.6 0.999999998869978
0.7 0.99999857290027
0.8 0.999897180595153
0.9 0.998283318978398
1 0.98817993034985
1.1 0.954499368567494
1.2 0.88238322312223
1.3 0.770109264089317
1.4 0.631738967436461
1.5 0.488426235243335
1.6 0.358220116360871
1.7 0.251152278540543
1.8 0.16961171159941
1.9 0.11109898777971
2 0.0710131882324527
2.1 0.0445256450697417
2.2 0.027506842417679
2.3 0.0168046929172315
2.4 0.0101836167889724
2.5 0.00613677052673833
2.6 0.00368492901741024
2.7 0.00220845719489648
2.8 0.00132282408579654
2.9 0.000792755323229866
3 0.000475750748254011
3.1 0.000286106476016124
3.2 0.00017251527884421
3.3 0.000104345369502878
3.4 6.33312409136222e-05
3.5 3.85817786129339e-05
3.6 2.35971162777515e-05
3.7 1.44916803196393e-05
3.8 8.93745841601401e-06
3.9 5.53582162532429e-06
4 3.44387841788585e-06
};
\addplot [very thick, gray]
table {%
0.5 0.999997197361275
0.6 0.999987034150312
0.7 0.999898006998428
0.8 0.999140092991866
0.9 0.994317198753636
1 0.975351268858724
1.1 0.935834144931911
1.2 0.88850748133696
1.3 0.828046832433802
1.4 0.731902233167783
1.5 0.605782597090896
1.6 0.4648545452047
1.7 0.347613324492979
1.8 0.2730375520785
1.9 0.244509428123888
2 0.18765384106241
2.1 0.104784184328446
2.2 0.0675078314195225
2.3 0.0580238652794709
2.4 0.0460997780139699
2.5 0.0298151732992806
2.6 0.0124195432356958
2.7 0.00310006737644943
2.8 0.00197348857244862
2.9 0.00271800791596973
3 0.0019284437304646
3.1 0.00167175130405535
3.2 0.00215986788078639
3.3 0.00129029469773878
3.4 0.000288223118917433
3.5 2.44326128959321e-05
3.6 1.60719462001404e-06
3.7 8.42960028800996e-07
3.8 7.64562320544616e-07
3.9 7.45606272487862e-07
4 7.43574129925876e-07
};
\addplot [very thick, black, dashed]
table {%
0.5 1
0.6 1
0.7 1
0.8 1
0.9 1
1 0.999999999348448
1.1 0.99999950235446
1.2 0.999973209728235
1.3 0.999598995569285
1.4 0.997158301080912
1.5 0.987738504067729
1.6 0.962905518012083
1.7 0.913879059405361
1.8 0.836569727613501
1.9 0.734465168099655
2 0.617381972771856
2.1 0.497529753222982
2.2 0.385599610401107
2.3 0.288575568937833
2.4 0.209429765332663
2.5 0.14799359805209
2.6 0.102200293054103
2.7 0.0691792751327623
2.8 0.0460033719080388
2.9 0.0300971283204075
3 0.0193911312382948
3.1 0.0123154370919198
3.2 0.00772044024779683
3.3 0.004785490293998
3.4 0.0029387118341998
3.5 0.00179149195992223
3.6 0.00108628571142166
3.7 0.000656315154883846
3.8 0.000395725169891836
3.9 0.000238431669817296
4 0.000143716534243055
};
\addplot [very thick, gray, dashed]
table {%
0.5 0.999999999999596
0.6 0.999999999989993
0.7 0.999999999872427
0.8 0.999999998347337
0.9 0.999999961134267
1 0.999999486545492
1.1 0.999994469367594
1.2 0.999897125319114
1.3 0.998824304624095
1.4 0.994376595243337
1.5 0.987105987362144
1.6 0.971866319884424
1.7 0.915861294402026
1.8 0.813075245772678
1.9 0.715433723141986
2 0.632609800573047
2.1 0.537950131261312
2.2 0.440676965324466
2.3 0.377930208610543
2.4 0.334950364261018
2.5 0.263379213862592
2.6 0.178248375168084
2.7 0.109216020871745
2.8 0.0757216209028111
2.9 0.0578156055573661
3 0.0311286832444155
3.1 0.0110906779439487
3.2 0.00902515279679174
3.3 0.0132127986952785
3.4 0.0105343445672704
3.5 0.005414720274106
3.6 0.00210133098404932
3.7 0.000475201399828539
3.8 5.44441595803258e-05
3.9 3.43763975983985e-06
4 2.00472181249124e-07
};
\addplot [very thick, black, dotted]
table {%
0.5 1
0.6 1
0.7 1
0.8 1
0.9 1
1 1
1.1 1
1.2 1
1.3 0.999999999999999
1.4 0.999999999377245
1.5 0.999999657092481
1.6 0.999985156187676
1.7 0.999809817862939
1.8 0.998773565728781
1.9 0.994896330962281
2 0.984383486717914
2.1 0.962047971359528
2.2 0.922923942475201
2.3 0.864322979544098
2.4 0.787151363442989
2.5 0.69581547694205
2.6 0.596926356773646
2.7 0.497553920246217
2.8 0.403757956805445
2.9 0.319768708556622
3 0.247822552515213
3.1 0.18844874034795
3.2 0.140960526053158
3.3 0.10396026119255
3.4 0.0757525703567505
3.5 0.0546306025901766
3.6 0.0390439069936283
3.7 0.0276755214076069
3.8 0.0194589062185967
3.9 0.0135605539107035
4 0.00934782588289695
};
\addplot [very thick, gray, dotted]
table {%
0.5 1
0.6 1
0.7 1
0.8 1
0.9 0.999999999999948
1 0.999999999987248
1.1 0.999999998852058
1.2 0.999999961984722
1.3 0.999999536596312
1.4 0.999997868480302
1.5 0.999992037510258
1.6 0.999853310691728
1.7 0.998250223094159
1.8 0.991929559410779
1.9 0.984006244856772
2 0.974663233884577
2.1 0.94141331796892
2.2 0.886655844973459
2.3 0.830810101938102
2.4 0.771199427176764
2.5 0.724813744956869
2.6 0.682867791034315
2.7 0.619584978607651
2.8 0.525237019488186
2.9 0.394702390000609
3 0.296021521248761
3.1 0.250470886972984
3.2 0.214808191058744
3.3 0.165272775211773
3.4 0.124783286078251
3.5 0.0952794542212789
3.6 0.0725304497720897
3.7 0.0404067426135432
3.8 0.0212791489713592
3.9 0.0213480622509047
4 0.0182532137874636
};
\end{axis}

\end{tikzpicture}
		\caption{Comparison of $\wangstamatiadis$ of \cref{eq:ws} (black) and the approximation of $\probabilityestcond{\collision}{\situationinitial}$ using \cref{eq:nadaraya watson} (gray) with $\simulationthreshold=0.2$.}
		\label{fig:ws comparison coarse nw}
	\end{subfigure}\hfill
	\begin{subfigure}{.49\linewidth}
		\centering
\begin{tikzpicture}

\definecolor{darkgray176}{RGB}{176,176,176}

\begin{axis}[
height=\figureheight,
scaled y ticks=false,
tick align=outside,
tick pos=left,
width=\figurewidth,
x grid style={darkgray176},
xlabel={$\ttcsymbol$ [\si{\second}]},
xmajorgrids,
xmin=0.5, xmax=4,
xtick style={color=black},
xticklabel style={align=center},
y grid style={darkgray176},
ylabel={Probability of collision},
ymajorgrids,
ymin=0, ymax=1,
ytick style={color=black},
yticklabel style={/pgf/number format/fixed,/pgf/number format/precision=3}
]
\addplot [very thick, black]
table {%
0.5 1
0.6 0.999999998869978
0.7 0.99999857290027
0.8 0.999897180595153
0.9 0.998283318978398
1 0.98817993034985
1.1 0.954499368567494
1.2 0.88238322312223
1.3 0.770109264089317
1.4 0.631738967436461
1.5 0.488426235243335
1.6 0.358220116360871
1.7 0.251152278540543
1.8 0.16961171159941
1.9 0.11109898777971
2 0.0710131882324527
2.1 0.0445256450697417
2.2 0.027506842417679
2.3 0.0168046929172315
2.4 0.0101836167889724
2.5 0.00613677052673833
2.6 0.00368492901741024
2.7 0.00220845719489648
2.8 0.00132282408579654
2.9 0.000792755323229866
3 0.000475750748254011
3.1 0.000286106476016124
3.2 0.00017251527884421
3.3 0.000104345369502878
3.4 6.33312409136222e-05
3.5 3.85817786129339e-05
3.6 2.35971162777515e-05
3.7 1.44916803196393e-05
3.8 8.93745841601401e-06
3.9 5.53582162532429e-06
4 3.44387841788585e-06
};
\addplot [very thick, gray]
table {%
0.5 0.999996357702958
0.6 0.999962696306387
0.7 0.99963563703153
0.8 0.997927114277664
0.9 0.992928312710559
1 0.979673117474703
1.1 0.939728412189211
1.2 0.860763259062648
1.3 0.762135641800424
1.4 0.645159977464581
1.5 0.513961930777097
1.6 0.399353905051341
1.7 0.299494987713676
1.8 0.216544926060255
1.9 0.148489443100922
2 0.0918225088994234
2.1 0.0516998384815958
2.2 0.0336747415960447
2.3 0.0273188588914856
2.4 0.0153109916100694
2.5 0.00513568819465852
2.6 0.00238872839023134
2.7 0.00163124145364462
2.8 0.000943667325736261
2.9 0.000388460801820716
3 7.81133955084528e-05
3.1 7.05226285914884e-06
3.2 1.51297601293087e-06
3.3 1.73124288382112e-06
3.4 1.6589426940256e-06
3.5 2.24757343119302e-06
3.6 2.95495881200575e-06
3.7 2.07496708426498e-06
3.8 1.04715162954753e-06
3.9 7.71249442091208e-07
4 7.44674926046043e-07
};
\addplot [very thick, black, dashed]
table {%
0.5 1
0.6 1
0.7 1
0.8 1
0.9 1
1 0.999999999348448
1.1 0.99999950235446
1.2 0.999973209728235
1.3 0.999598995569285
1.4 0.997158301080912
1.5 0.987738504067729
1.6 0.962905518012083
1.7 0.913879059405361
1.8 0.836569727613501
1.9 0.734465168099655
2 0.617381972771856
2.1 0.497529753222982
2.2 0.385599610401107
2.3 0.288575568937833
2.4 0.209429765332663
2.5 0.14799359805209
2.6 0.102200293054103
2.7 0.0691792751327623
2.8 0.0460033719080388
2.9 0.0300971283204075
3 0.0193911312382948
3.1 0.0123154370919198
3.2 0.00772044024779683
3.3 0.004785490293998
3.4 0.0029387118341998
3.5 0.00179149195992223
3.6 0.00108628571142166
3.7 0.000656315154883846
3.8 0.000395725169891836
3.9 0.000238431669817296
4 0.000143716534243055
};
\addplot [very thick, gray, dashed]
table {%
0.5 0.999999999999998
0.6 0.999999999999501
0.7 0.999999999892155
0.8 0.999999990178621
0.9 0.999999653122776
1 0.999994836331739
1.1 0.999957451212964
1.2 0.999712599865538
1.3 0.998388670487841
1.4 0.993498588899135
1.5 0.981315794849655
1.6 0.959877117418424
1.7 0.926298179058338
1.8 0.866920247970333
1.9 0.769991233125085
2 0.646842550920723
2.1 0.526302980485779
2.2 0.419728905535399
2.3 0.305893491535558
2.4 0.189772603481342
2.5 0.107207149354285
2.6 0.0823735011290323
2.7 0.0647890548948959
2.8 0.0335647897507074
2.9 0.0163003590711636
3 0.0121499118863092
3.1 0.00634859496137853
3.2 0.0015498036986122
3.3 0.000331618985688212
3.4 0.000283198212981562
3.5 0.000825381651738068
3.6 0.00292796063309384
3.7 0.00462990208004172
3.8 0.0027878652597661
3.9 0.000655234031165362
4 7.24669082936768e-05
};
\addplot [very thick, black, dotted]
table {%
0.5 1
0.6 1
0.7 1
0.8 1
0.9 1
1 1
1.1 1
1.2 1
1.3 0.999999999999999
1.4 0.999999999377245
1.5 0.999999657092481
1.6 0.999985156187676
1.7 0.999809817862939
1.8 0.998773565728781
1.9 0.994896330962281
2 0.984383486717914
2.1 0.962047971359528
2.2 0.922923942475201
2.3 0.864322979544098
2.4 0.787151363442989
2.5 0.69581547694205
2.6 0.596926356773646
2.7 0.497553920246217
2.8 0.403757956805445
2.9 0.319768708556622
3 0.247822552515213
3.1 0.18844874034795
3.2 0.140960526053158
3.3 0.10396026119255
3.4 0.0757525703567505
3.5 0.0546306025901766
3.6 0.0390439069936283
3.7 0.0276755214076069
3.8 0.0194589062185967
3.9 0.0135605539107035
4 0.00934782588289695
};
\addplot [very thick, gray, dotted]
table {%
0.5 1
0.6 1
0.7 1
0.8 1
0.9 1
1 0.999999999999976
1.1 0.999999999994165
1.2 0.999999999475009
1.3 0.999999982422203
1.4 0.999999762609874
1.5 0.999997531237392
1.6 0.999956667248916
1.7 0.999517958506481
1.8 0.997759803167201
1.9 0.995280765617134
2 0.991860522133354
2.1 0.977586931997455
2.2 0.940266867968824
2.3 0.882337076150675
2.4 0.807225974633966
2.5 0.71637882034375
2.6 0.629624071136963
2.7 0.542121866006725
2.8 0.448364598356913
2.9 0.340394302858719
3 0.232741542357195
3.1 0.173641193371774
3.2 0.143688781316322
3.3 0.107625353046941
3.4 0.0749302276804413
3.5 0.0518879275141784
3.6 0.0256319766595491
3.7 0.00822147315542865
3.8 0.00569713781274151
3.9 0.00734092953608462
4 0.00717744281143517
};
\end{axis}

\end{tikzpicture}
		\caption{Comparison of $\wangstamatiadis$ of \cref{eq:ws} (black) and the approximation of $\probabilityestcond{\collision}{\situationinitial}$ using \cref{eq:nadaraya watson} (gray) with $\simulationthreshold=0.02$.}
		\label{fig:ws comparison fine nw}
	\end{subfigure}
	\caption{Comparison of $\wangstamatiadis$ of \cref{eq:ws} (black lines) and $\probabilityestcond{\collision}{\situationinitial}$ of \cref{eq:estimate probability of collision} (a,b) or the approximation of $\probabilityestcond{\collision}{\situationinitial}$ using \cref{eq:nadaraya watson} (c,d) (gray lines) for a speed difference of $\speeddifferencesymbol=\SI{10}{\meter\per\second}$ (solid lines), $\speeddifferencesymbol=\SI{20}{\meter\per\second}$ (dashed lines), and $\speeddifferencesymbol=\SI{30}{\meter\per\second}$ (dotted lines).
		Here, $\probabilityestcond{\collision}{\situationinitial}$ is based on the same underlying assumptions as $\wangstamatiadis$, see \cref{sec:wang stamatiadis explanation}.
		The influence of the parameter $\simulationthreshold$, which determines the number of simulations to estimate $\probabilitycond{\collision}{\situationinitial}$, is illustrated by using different values.}
	\label{fig:ws comparison}
\end{figure}

\subsection{Developing an SSM for longitudinal interactions}
\label{sec:ngsim metric}

To further illustrate the \ac{ourmethod} method, we apply it to derive \iac{ssm} that calculates the risk of a crash in a longitudinal interaction between two vehicles.
The \ac{ssm} is based on the \ac{ngsim} data set \autocite{alexiadis2004next}.
The \ac{ngsim} data set contains vehicle trajectories obtained from video footage of cameras that were located at several motorways in the U.S.A. 
The derived \ac{ssm} estimates the risk of a crash of the ego vehicle with its leading vehicle.
To describe the initial situation at time $\time$, $\situationinitialdim=4$ parameters are used:
\begin{itemize}
	\item the speed of the leading vehicle ($\speedlead{\time}$);
	\item the acceleration of the leading vehicle ($\accelerationlead{\time}$);
	\item the speed of the ego vehicle ($\speedego{\time}$); and
	\item the log of the gap between the leading vehicle and the ego vehicle\footnote{Note that the log is used, such that there are, relatively speaking, more simulations performed with a small initial gap, cf.\ \cref{eq:design points distance}.} $\lnsymbol \gap{\time}$.
\end{itemize}
Thus, we have:
\begin{equation}
	\label{eq:initial situatino ngsim}
	\situationinitial\transpose(\time) = \begin{bmatrix}
		\speedlead{\time} & \accelerationlead{\time} & \speedego{\time} & \lnsymbol \gap{\time}
	\end{bmatrix}.
\end{equation}

The speed of the leading vehicle at $\situationfuturehorizon=50$ instances, each $\situationfuturetimestep=\SI{0.1}{\second}$ apart, describes the future situation:
\begin{equation}
	\label{eq:example future situation}
	\situationfuture\transpose(\time) = \begin{bmatrix}
		\speedlead{\time+\situationfuturetimestep} & \cdots & \speedlead{\time+\situationfuturehorizon\situationfuturetimestep}
	\end{bmatrix}.
\end{equation}
It is assumed that $\situationfuture(\time)$ depends on $\speedlead{\time}$ and $\accelerationlead{\time}$. 
To model this dependency with a single kernel density estimator would give us \iac{pdf} with $\situationfuturehorizon+2$ dimensions.
To reduce the dimensionality, we use \iac{svd} as described in \cref{sec:parameter reduction} with\footnote{Note that because we assume that $\situationfuture(\time)$ depends on 2 parameters of $\situationinitial(\time)$, i.e., $\speedlead{\time}$ and $\accelerationlead{\time}$, we need to choose $\dimension$ such that $2 < \dimension < \situationfuturehorizon+2$.} $\dimension=4$.
In total, 18182 longitudinal interactions between two vehicles have been analyzed.
\cstart Here, a longitudinal interaction between a leading and following vehicle within the same lane always begins when the \ac{thw} is less than or equal to \SI{2}{\second}, or when $\gapsymbol$ is less than or equal to \SI{20}{\meter}. 
The interaction ceases when both the \ac{thw} exceeds \SI{4}{\second} and $\gapsymbol$ becomes greater than \SI{40}{\meter}, or when both vehicles are no longer the closest pair in the same lane. 
This scheme is commonly referred to as a hysteresis loop. \cend
For each second of an interaction, we extract an ``initial situation'' $\situationinitialinstance{\situationindex}$ \cstart according to \cref{eq:initial situatino ngsim} \cend and a corresponding ``future situation'' $\situationfutureinstance{\situationindex}$ \cstart according to \cref{eq:example future situation}\cend. 
This leads to $\situationnumberof=469453$ data points.
Based on Silverman's rule of thumb \autocite{silverman1986density}, we use a bandwidth matrix $\bandwidthmatrix=\bandwidth^2\identitymatrix{4}$ for the \ac{kde} with $\bandwidth\approx 0.186$ and $\identitymatrix{4}$ denoting the 4-by-4 identity matrix.

To demonstrate the sampling from the estimated density of the reduced parameter vector subject to a linear constraint such as \cref{eq:linear constraint}, the plots in \cref{fig:speed profiles} show 50 different future situations in the form of \cref{eq:example future situation}.
\Cref{fig:speed profiles accelerating} assumes an initial situation with $\speedleadsymbol=\SI{15}{\meter\per\second}$ and $\accelerationleadsymbol=\SI{1}{\meter\per\second\squared}$ and \cref{fig:speed profiles decelerating} assumes an initial situation with $\speedleadsymbol=\SI{15}{\meter\per\second}$ and $\accelerationleadsymbol=\SI{-1}{\meter\per\second\squared}$.
Note that the same \ac{pdf} is used to produce the lines in \cref{fig:speed profiles}; the only difference between \cref{fig:speed profiles accelerating} and \cref{fig:speed profiles decelerating} is a different linear constraint (based on $\speedleadsymbol$ and $\accelerationleadsymbol$) on the generated samples.
In case a simulation run is longer than \SI{5}{\second}, the speed of the leading vehicle is assumed to remain constant after these \SI{5}{\second}.
Note that a simulation run is rarely longer than \SI{5}{\second}, so this assumption does not have a significant effect on the results.

\setlength{\figurewidth}{.49\linewidth}
\setlength{\figureheight}{.7\figurewidth}
\begin{figure}
	\centering
	\begin{subfigure}{.49\linewidth}
		\centering
		\input{figs/speed_profiles_a.tikz}
		\caption{Initial situation: $\speedleadsymbol=\SI{15}{\meter\per\second}$ and $\accelerationleadsymbol=\SI{1}{\meter\per\second\squared}$.}
		\label{fig:speed profiles accelerating}
	\end{subfigure}
	\begin{subfigure}{.49\linewidth}
		\centering
		\input{figs/speed_profiles_b.tikz}
		\caption{Initial situation: $\speedleadsymbol=\SI{15}{\meter\per\second}$ and $\accelerationleadsymbol=\SI{-1}{\meter\per\second\squared}$.}
		\label{fig:speed profiles decelerating}
	\end{subfigure}
	\caption{50 potential future situations samples from the \ac{kde} that is constructed using data from the \ac{ngsim} data set.}
	\label{fig:speed profiles}
\end{figure}

To estimate $\probabilitycond{\collision}{\situationinitial}$ (\cref{sec:estimate collision}), we use the \ac{idmplus} \autocite{schakel2010effects} for modeling the ego vehicle driver behavior and response.
In addition to \ac{idmplus}, we assume that the driver has a reaction time that is similarly distributed as $\timereact$ in \cref{sec:wang stamatiadis explanation} and that the \ac{madr} is similarly distributed as $\accelerationmax$ in \cref{sec:wang stamatiadis explanation}.
The simulation result $\simulationresult$ is defined according to \cref{eq:simulation result}.
The minimum number of simulations to estimate $\probabilitycond{\collision}{\situationinitial}$ is set to 10 and this number is further increased until the condition in \cref{eq:condition stop simulations} with $\simulationthreshold=0.1$ is met.

To calculate $\probabilitycond{\collision}{\situationinitial}$ using \cref{eq:nadaraya watson}, we create a grid of points  $\left\{\situationinitialinstance{\situationindexdesign}'\right\}_{\situationindexdesign=1}^{\numberofdesignpoints}$ using the method explained in \cref{sec:final metric calculation}.
For $\weightmatrix$, we use a diagonal matrix with diagonal elements: $0.25$, $4$, $0.25$, and $0.25$, which is a trade-off between keeping many points such that the estimation in \cref{eq:nadaraya watson} is accurate while also keeping the total number of points for which $\probabilitycond{\collision}{\situationinitial}$ is estimated manageable.
With this choice of $\weightmatrix$, we have $\numberofdesignpoints=10129$.
For the regression of \cref{eq:nadaraya watson}, we use $\bandwidthnw=\weightmatrix^{-1}$.

\subsection{Analyzing the SSMs for longitudinal interactions}
\label{sec:analyzing ngsim metric}

The heat maps in \cref{fig:heatmaps} show how the developed \ac{ssm} depends on the input variables $\speedleadsymbol$ and $\gapsymbol$. 
The other two parameters, $\speedegosymbol$ and $\accelerationleadsymbol$, are fixed for each heat map.
The heat maps show that the estimated crash probability is practically 0 if both $\speedleadsymbol$ and $\gapsymbol$ are large.
This seems reasonable, because in that case, the ego vehicle is at a safe distance from its leading vehicle while the approaching speed is small. 
In addition, for a fixed $\speedegosymbol$, we see that the crash risk increases as the difference in speed increases, as is expected.
The same applies for a decreasing distance between the two vehicles.
For small values of $\speedleadsymbol$ and $\gapsymbol$, the estimated crash probability is practically 1.
The left and center heat maps of \cref{fig:heatmaps} show that for a higher speed of the ego vehicle, the crash probability is estimated to be higher.
Similarly, the right and center heat maps of \cref{fig:heatmaps} show that for a lower initial acceleration of the leading vehicle, the crash probability is estimated to be higher.

\setlength{\figurewidth}{.3\linewidth}
\setlength{\figureheight}{0.8\figurewidth}
\begin{figure}
	\centering
	\input{figs/heatmaps_edited.tikz}
	\caption{Heat maps of the \ac{ssm} described in \cref{sec:ngsim metric} as a function of the speed of the leading vehicle ($\speedleadsymbol$) and the gap between the ego vehicle and the leading vehicle ($\gapsymbol$).
		For each heat map, the other two input parameters are fixed at $\speedegosymbol=\SI{25}{\meter\per\second}$ (left) or $\speedegosymbol=\SI{20}{\meter\per\second}$ (center and right) and $\accelerationleadsymbol=\SI{0}{\meter\per\second\squared}$ (left and center) or $\accelerationleadsymbol=\SI{-1}{\meter\per\second\squared}$ (right).
		The estimated crash probability ranges from 0 (white) to 1 (black).}
	\label{fig:heatmaps}
\end{figure}

In \cref{fig:scenarios}, the evaluations of the measure described in \cref{sec:ngsim metric} are shown for 3 different scenarios. 
Each of the 3 scenarios considers an ego vehicle and a leading vehicle driving in front of the ego vehicle.
Both vehicles are driving in the same direction and in the same lane. 
For comparison, the right plots also include the evaluations of $\wangstamatiadis$ of \cref{eq:ws}.

\setlength{\figurewidth}{.45\linewidth}
\setlength{\figureheight}{0.6\figurewidth}
\begin{figure}
	\centering
\begin{tikzpicture}

\definecolor{darkgray176}{RGB}{176,176,176}

\begin{axis}[
height=\figureheight,
scaled y ticks=false,
tick align=outside,
tick pos=left,
width=\figurewidth,
x grid style={darkgray176},
xlabel={Time [\si{\second}]},
xmajorgrids,
xmin=0, xmax=12,
xtick style={color=black},
xticklabel style={align=center},
y grid style={darkgray176},
ylabel=\textcolor{black}{Speed [\si{\meter\per\second}]},
ymajorgrids,
ymin=7.2, ymax=24.8,
ytick style={color=black},
yticklabel style={/pgf/number format/fixed,/pgf/number format/precision=3}
]
\addplot [very thick, black]
table {%
0 24
0.1 24
0.2 24
0.3 24
0.4 24
0.5 24
0.6 24
0.7 24
0.8 24
0.9 24
1 24
1.1 24
1.2 24
1.3 24
1.4 24
1.5 24
1.6 24
1.7 24
1.8 24
1.9 24
2 24
2.1 23.975338669865
2.2 23.9015067247611
2.3 23.7789593631814
2.4 23.6084521303612
2.5 23.3910362600903
2.6 23.1280521935069
2.7 22.8211213148327
2.8 22.4721359549996
2.9 22.0832477248002
3 21.6568542494924
3.1 21.1955843866415
3.2 20.7022820183398
3.3 20.1799885177276
3.4 19.6319239979164
3.5 19.0614674589207
3.6 18.4721359549996
3.7 17.8675629108472
3.8 17.2514757203218
3.9 16.6276727658228
4 16
4.1 15.3723272341772
4.2 14.7485242796782
4.3 14.1324370891528
4.4 13.5278640450004
4.5 12.9385325410793
4.6 12.3680760020836
4.7 11.8200114822724
4.8 11.2977179816602
4.9 10.8044156133585
5 10.3431457505076
5.1 9.91675227519975
5.2 9.52786404500042
5.3 9.17887868516726
5.4 8.87194780649306
5.5 8.60896373990971
5.6 8.39154786963877
5.7 8.22104063681859
5.8 8.0984932752389
5.9 8.02466133013498
6 8
6.1 8.00308266626687
6.2 8.01231165940486
6.3 8.02763007960232
6.4 8.04894348370485
6.5 8.07612046748871
6.6 8.10899347581163
6.7 8.14735983564591
6.8 8.19098300562505
6.9 8.23959403439997
7 8.29289321881345
7.1 8.35055195166982
7.2 8.41221474770753
7.3 8.47750143528405
7.4 8.54600950026045
7.5 8.61731656763491
7.6 8.69098300562505
7.7 8.76655463614409
7.8 8.84356553495977
7.9 8.92154090427216
8 9
8.1 9.07845909572784
8.2 9.15643446504023
8.3 9.23344536385591
8.4 9.30901699437495
8.5 9.38268343236509
8.6 9.45399049973955
8.7 9.52249856471595
8.8 9.58778525229247
8.9 9.64944804833019
9 9.70710678118655
9.1 9.76040596560003
9.2 9.80901699437495
9.3 9.85264016435409
9.4 9.89100652418837
9.5 9.92387953251129
9.6 9.95105651629515
9.7 9.97236992039768
9.8 9.98768834059514
9.9 9.99691733373313
10 10
10.1 10
10.2 10
10.3 10
10.4 10
10.5 10
10.6 10
10.7 10
10.8 10
10.9 10
11 10
11.1 10
11.2 10
11.3 10
11.4 10
11.5 10
11.6 10
11.7 10
11.8 10
11.9 10
12 10
};
\addplot [very thick, black, dashed]
table {%
0 20
0.1 20
0.2 20
0.3 20
0.4 20
0.5 20
0.6 20
0.7 20
0.8 20
0.9 20
1 20
1.1 20
1.2 20
1.3 20
1.4 20
1.5 20
1.6 20
1.7 20
1.8 20
1.9 20
2 20
2.1 20
2.2 20
2.3 20
2.4 20
2.5 20
2.6 20
2.7 20
2.8 20
2.9 20
3 20
3.1 19.9778098230154
3.2 19.9114362536434
3.3 19.8014684283847
3.4 19.6488824294413
3.5 19.4550326209418
3.6 19.2216396275101
3.7 18.9507750618785
3.8 18.6448431371071
3.9 18.3065593266183
4 17.9389262614624
4.1 17.5452070787519
4.2 17.1288964578254
4.3 16.6936896012265
4.4 16.2434494358243
4.5 15.7821723252012
4.6 15.3139525976466
4.7 14.8429462046094
4.8 14.3733338321785
4.9 13.9092837930173
5 13.4549150281253
5.1 13.0142605468261
5.2 12.5912316294914
5.3 12.1895831107393
5.4 11.8128800512565
5.5 11.4644660940673
5.6 11.1474337861211
5.7 10.8645971286272
5.8 10.6184665997807
5.9 10.4112268715801
6 10.2447174185242
6.1 10.1204161903063
6.2 10.0394264934276
6.3 10.0024671981713
6.4 10
6.5 10
6.6 10
6.7 10
6.8 10
6.9 10
7 10
7.1 10
7.2 10
7.3 10
7.4 10
7.5 10
7.6 10
7.7 10
7.8 10
7.9 10
8 10
8.1 10
8.2 10
8.3 10
8.4 10
8.5 10
8.6 10
8.7 10
8.8 10
8.9 10
9 10
9.1 10
9.2 10
9.3 10
9.4 10
9.5 10
9.6 10
9.7 10
9.8 10
9.9 10
10 10
10.1 10
10.2 10
10.3 10
10.4 10
10.5 10
10.6 10
10.7 10
10.8 10
10.9 10
11 10
11.1 10
11.2 10
11.3 10
11.4 10
11.5 10
11.6 10
11.7 10
11.8 10
11.9 10
12 10
};
\end{axis}

\begin{axis}[
axis y line=right,
height=\figureheight,
scaled y ticks=false,
tick align=outside,
width=\figurewidth,
x grid style={darkgray176},
xmin=0, xmax=12,
xtick pos=left,
xtick style={color=black},
xticklabel style={align=center},
y grid style={darkgray176},
ylabel=\textcolor{gray}{Distance [\si{\meter}]},
ymin=27.2, ymax=44.8,
ytick pos=right,
ytick style={color=black},
yticklabel style={/pgf/number format/fixed,/pgf/number format/precision=3},
yticklabel style={anchor=west}
]
\addplot [very thick, gray, dotted]
table {%
0 39.96
0.1 39.56
0.2 39.16
0.3 38.76
0.4 38.36
0.5 37.96
0.6 37.56
0.7 37.16
0.8 36.76
0.9 36.36
1 35.96
1.1 35.56
1.2 35.16
1.3 34.76
1.4 34.36
1.5 33.96
1.6 33.56
1.7 33.16
1.8 32.76
1.9 32.36
2 31.96
2.1 31.5609496281647
2.2 31.1670722808844
2.3 30.7832624759959
2.4 30.4143526615897
2.5 30.0650834223601
2.6 29.7400742523241
2.7 29.4437950741051
2.8 29.1805386803732
2.9 28.9543942673507
3 28.7692222235582
3.1 28.627775791076
3.2 28.529590084631
3.3 28.4733134771088
3.4 28.4573604670254
3.5 28.4799331339793
3.6 28.5390435825157
3.7 28.6325371352064
3.8 28.7581160348726
3.9 28.9133634173509
4 29.095767319958
4.1 29.3027444967952
4.2 29.5316638201412
4.3 29.7798690573084
4.4 30.0447008243527
4.5 30.3235175317815
4.6 30.6137151527394
4.7 30.9127456608884
4.8 31.2181340031529
4.9 31.5274934914745
5 31.8385395175075
5.1 32.1491015145749
5.2 32.4571331119845
5.3 32.7607204477502
5.4 33.0580886266708
5.5 33.3476063313598
5.6 33.6277886140007
5.7 33.8972979161074
5.8 34.1549433822126
5.9 34.3996785510051
6 34.6305975238114
6.1 34.8477606499648
6.2 35.0542216307064
6.3 35.2537425454437
6.4 35.4498725565523
6.5 35.643531211456
6.6 35.8341573577105
6.7 36.0211922549879
6.8 36.2040993042044
6.9 36.3823673558322
7 36.5555138613062
7.1 36.7230878480966
7.2 36.8846727001042
7.3 37.0398887262372
7.4 37.1883955013333
7.5 37.329893964994
7.6 37.4641282653912
7.7 37.5908873366788
7.8 37.7100062002841
7.9 37.8213669820552
8 37.9248996389948
8.1 38.0205823910979
8.2 38.1084418556337
8.3 38.1885528830403
8.4 38.2610380954475
8.5 38.3260671306704
8.6 38.3838555963359
8.7 38.43466374059
8.8 38.4787948475812
8.9 38.5165933676133
9 38.5484427934944
9.1 38.5747632961762
9.2 38.5960091342606
9.3 38.6126658533443
9.4 38.6252472924686
9.5 38.6342924161309
9.6 38.6403619913893
9.7 38.6440351305476
9.8 38.6459057207345
9.9 38.6465787623931
10 38.646666639251
10.1 38.646666639251
10.2 38.646666639251
10.3 38.646666639251
10.4 38.646666639251
10.5 38.646666639251
10.6 38.646666639251
10.7 38.646666639251
10.8 38.646666639251
10.9 38.646666639251
11 38.646666639251
11.1 38.646666639251
11.2 38.646666639251
11.3 38.646666639251
11.4 38.646666639251
11.5 38.646666639251
11.6 38.646666639251
11.7 38.646666639251
11.8 38.646666639251
11.9 38.646666639251
12 38.646666639251
};
\end{axis}

\end{tikzpicture}
\begin{tikzpicture}

\definecolor{darkgray176}{RGB}{176,176,176}

\begin{axis}[
height=\figureheight,
scaled y ticks=false,
tick align=outside,
tick pos=left,
width=\figurewidth,
x grid style={darkgray176},
xlabel={Time [\si{\second}]},
xmajorgrids,
xmin=0, xmax=12,
xtick style={color=black},
xticklabel style={align=center},
y grid style={darkgray176},
ylabel={Probability of a crash},
ymajorgrids,
ymin=0, ymax=1,
ytick style={color=black},
yticklabel style={/pgf/number format/fixed,/pgf/number format/precision=3}
]
\addplot [very thick, gray]
table {%
0 4.44089209850063e-16
0.1 4.44089209850063e-16
0.2 4.44089209850063e-16
0.3 7.7715611723761e-16
0.4 9.99200722162641e-16
0.5 1.33226762955019e-15
0.6 1.66533453693773e-15
0.7 2.22044604925031e-15
0.8 3.33066907387547e-15
0.9 4.32986979603811e-15
1 6.10622663543836e-15
1.1 8.21565038222616e-15
1.2 1.11022302462516e-14
1.3 1.50990331349021e-14
1.4 2.05391259555654e-14
1.5 2.80886425230165e-14
1.6 3.85247389544929e-14
1.7 5.26245713672324e-14
1.8 7.24975635080227e-14
1.9 9.96980276113391e-14
2 1.37889699658444e-13
2.1 1.61648472385423e-13
2.2 1.35780275911657e-13
2.3 7.97140131680862e-14
2.4 3.15303338993544e-14
2.5 7.99360577730113e-15
2.6 9.99200722162641e-16
2.7 0
2.8 0
2.9 0
3 0
3.1 0
3.2 0
3.3 0
3.4 0
3.5 0
3.6 0
3.7 0
3.8 0
3.9 0
4 0
4.1 0
4.2 0
4.3 0
4.4 0
4.5 0
4.6 0
4.7 0
4.8 0
4.9 0
5 0
5.1 0
5.2 0
5.3 0
5.4 0
5.5 0
5.6 0
5.7 0
5.8 0
5.9 0
6 0
6.1 0
6.2 0
6.3 0
6.4 0
6.5 0
6.6 0
6.7 0
6.8 0
6.9 0
7 0
7.1 0
7.2 0
7.3 0
7.4 0
7.5 0
7.6 0
7.7 0
7.8 0
7.9 0
8 0
8.1 0
8.2 0
8.3 0
8.4 0
8.5 0
8.6 0
8.7 0
8.8 0
8.9 0
9 0
9.1 0
9.2 0
9.3 0
9.4 0
9.5 0
9.6 0
9.7 0
9.8 0
9.9 0
10 0
10.1 0
10.2 0
10.3 0
10.4 0
10.5 0
10.6 0
10.7 0
10.8 0
10.9 0
11 0
11.1 0
11.2 0
11.3 0
11.4 0
11.5 0
11.6 0
11.7 0
11.8 0
11.9 0
12 0
};
\addplot [very thick, black]
table {%
0 5.98598443864919e-05
0.1 5.97315200644449e-05
0.2 5.96685792819155e-05
0.3 5.96774730194659e-05
0.4 5.97654703311494e-05
0.5 5.99407296333559e-05
0.6 6.02123736923254e-05
0.7 6.05905680130037e-05
0.8 6.10866022024672e-05
0.9 6.17129737323837e-05
1 6.24834733570863e-05
1.1 6.34132712580455e-05
1.2 6.45190027844299e-05
1.3 6.58188524474615e-05
1.4 6.73326346100528e-05
1.5 6.90818691021935e-05
1.6 7.10898497995121e-05
1.7 7.33817040440434e-05
1.8 7.59844406836249e-05
1.9 7.89269844856417e-05
2 8.22401947735142e-05
2.1 8.57232350662187e-05
2.2 8.90210493045708e-05
2.3 9.1902873464813e-05
2.4 9.41181413431408e-05
2.5 9.54492160687657e-05
2.6 9.57965769964997e-05
2.7 9.52906599224754e-05
2.8 9.44062045272763e-05
2.9 9.40287856706681e-05
3 9.53982953196404e-05
3.1 0.000253545693576774
3.2 0.000501187987125852
3.3 0.000717187866048598
3.4 0.000916875582591368
3.5 0.00111363205237602
3.6 0.0012321562688624
3.7 0.00127058859518567
3.8 0.00113276205349576
3.9 0.000885706437739368
4 0.000860887778127544
4.1 0.00130190829680953
4.2 0.00212687785418121
4.3 0.00291325994171351
4.4 0.0032720981887869
4.5 0.00316509684109677
4.6 0.00283291489002934
4.7 0.00258346393463606
4.8 0.00261688357602838
4.9 0.00294839323006407
5 0.00344286729706845
5.1 0.00393421511619057
5.2 0.00450662419330526
5.3 0.00624602163819999
5.4 0.0117793846143548
5.5 0.0227217815175308
5.6 0.0370772067310604
5.7 0.0486320423792221
5.8 0.0499833332609049
5.9 0.0413462032322658
6 0.0292189324282003
6.1 0.0187061397542497
6.2 0.0113414907078916
6.3 0.00623462172965389
6.4 0.00495374106365929
6.5 0.00504891305130941
6.6 0.00515166206765486
6.7 0.00526186021546665
6.8 0.00537930371461506
6.9 0.00550370329195404
7 0.00563467534147094
7.1 0.00577173456958721
7.2 0.00591428888765529
7.3 0.00606163730929337
7.4 0.00621297154273164
7.5 0.00636738183075658
7.6 0.00652386738274954
7.7 0.00668135147267934
7.8 0.00683870096071749
7.9 0.0069947496597396
8 0.00714832464283744
8.1 0.00729827430811456
8.2 0.00744349681453329
8.3 0.00758296740283604
8.4 0.00771576313336079
8.5 0.00784108370954588
8.6 0.00795826730006902
8.7 0.00806680059987088
8.8 0.00816632274820726
8.9 0.00825662311385442
9 0.00833763332814067
9.1 0.00840941426542051
9.2 0.00847213891623924
9.3 0.00852607225907285
9.4 0.00857154931042481
9.5 0.00860895252700678
9.6 0.00863868966071457
9.7 0.00866117304363049
9.8 0.00867680112371782
9.9 0.00868594289831196
10 0.00868892571523147
10.1 0.00868892571523147
10.2 0.00868892571523147
10.3 0.00868892571523147
10.4 0.00868892571523147
10.5 0.00868892571523147
10.6 0.00868892571523147
10.7 0.00868892571523147
10.8 0.00868892571523147
10.9 0.00868892571523147
11 0.00868892571523147
11.1 0.00868892571523147
11.2 0.00868892571523147
11.3 0.00868892571523147
11.4 0.00868892571523147
11.5 0.00868892571523147
11.6 0.00868892571523147
11.7 0.00868892571523147
11.8 0.00868892571523147
11.9 0.00868892571523147
12 0.00868892571523147
};
\end{axis}

\end{tikzpicture}\\
\begin{tikzpicture}

\definecolor{darkgray176}{RGB}{176,176,176}

\begin{axis}[
height=\figureheight,
scaled y ticks=false,
tick align=outside,
tick pos=left,
width=\figurewidth,
x grid style={darkgray176},
xlabel={Time [\si{\second}]},
xmajorgrids,
xmin=0, xmax=12,
xtick style={color=black},
xticklabel style={align=center},
y grid style={darkgray176},
ylabel=\textcolor{black}{Speed [\si{\meter\per\second}]},
ymajorgrids,
ymin=7.2, ymax=24.8,
ytick style={color=black},
yticklabel style={/pgf/number format/fixed,/pgf/number format/precision=3}
]
\addplot [very thick, black]
table {%
0 24
0.1 24
0.2 24
0.3 24
0.4 24
0.5 24
0.6 24
0.7 24
0.8 24
0.9 24
1 24
1.1 24
1.2 24
1.3 24
1.4 24
1.5 24
1.6 24
1.7 24
1.8 24
1.9 24
2 24
2.1 24
2.2 24
2.3 24
2.4 24
2.5 24
2.6 24
2.7 24
2.8 24
2.9 24
3 24
3.1 24
3.2 24
3.3 24
3.4 24
3.5 24
3.6 24
3.7 24
3.8 24
3.9 24
4 24
4.1 23.975338669865
4.2 23.9015067247611
4.3 23.7789593631814
4.4 23.6084521303612
4.5 23.3910362600903
4.6 23.1280521935069
4.7 22.8211213148327
4.8 22.4721359549996
4.9 22.0832477248002
5 21.6568542494924
5.1 21.1955843866415
5.2 20.7022820183398
5.3 20.1799885177276
5.4 19.6319239979164
5.5 19.0614674589207
5.6 18.4721359549996
5.7 17.8675629108472
5.8 17.2514757203218
5.9 16.6276727658228
6 16
6.1 15.3723272341772
6.2 14.7485242796782
6.3 14.1324370891528
6.4 13.5278640450004
6.5 12.9385325410793
6.6 12.3680760020836
6.7 11.8200114822724
6.8 11.2977179816602
6.9 10.8044156133585
7 10.3431457505076
7.1 9.91675227519975
7.2 9.52786404500042
7.3 9.17887868516726
7.4 8.87194780649306
7.5 8.60896373990971
7.6 8.39154786963877
7.7 8.22104063681859
7.8 8.0984932752389
7.9 8.02466133013498
8 8
8.1 8.00308266626687
8.2 8.01231165940486
8.3 8.02763007960232
8.4 8.04894348370485
8.5 8.07612046748871
8.6 8.10899347581163
8.7 8.14735983564591
8.8 8.19098300562505
8.9 8.23959403439997
9 8.29289321881345
9.1 8.35055195166981
9.2 8.41221474770753
9.3 8.47750143528405
9.4 8.54600950026045
9.5 8.61731656763491
9.6 8.69098300562505
9.7 8.7665546361441
9.8 8.84356553495977
9.9 8.92154090427216
10 9
10.1 9.07845909572784
10.2 9.15643446504023
10.3 9.23344536385591
10.4 9.30901699437495
10.5 9.38268343236509
10.6 9.45399049973955
10.7 9.52249856471595
10.8 9.58778525229247
10.9 9.64944804833019
11 9.70710678118655
11.1 9.76040596560003
11.2 9.80901699437495
11.3 9.85264016435409
11.4 9.89100652418837
11.5 9.92387953251129
11.6 9.95105651629515
11.7 9.97236992039768
11.8 9.98768834059514
11.9 9.99691733373313
12 10
};
\addplot [very thick, black, dashed]
table {%
0 20
0.1 20
0.2 20
0.3 20
0.4 20
0.5 20
0.6 20
0.7 20
0.8 20
0.9 20
1 20
1.1 20
1.2 20
1.3 20
1.4 20
1.5 20
1.6 20
1.7 20
1.8 20
1.9 20
2 20
2.1 20
2.2 20
2.3 20
2.4 20
2.5 20
2.6 20
2.7 20
2.8 20
2.9 20
3 20
3.1 19.9778098230154
3.2 19.9114362536434
3.3 19.8014684283847
3.4 19.6488824294413
3.5 19.4550326209418
3.6 19.2216396275101
3.7 18.9507750618785
3.8 18.6448431371071
3.9 18.3065593266183
4 17.9389262614624
4.1 17.5452070787519
4.2 17.1288964578254
4.3 16.6936896012265
4.4 16.2434494358243
4.5 15.7821723252012
4.6 15.3139525976466
4.7 14.8429462046094
4.8 14.3733338321785
4.9 13.9092837930173
5 13.4549150281253
5.1 13.0142605468261
5.2 12.5912316294914
5.3 12.1895831107393
5.4 11.8128800512565
5.5 11.4644660940673
5.6 11.1474337861211
5.7 10.8645971286272
5.8 10.6184665997807
5.9 10.4112268715801
6 10.2447174185242
6.1 10.1204161903063
6.2 10.0394264934276
6.3 10.0024671981713
6.4 10
6.5 10
6.6 10
6.7 10
6.8 10
6.9 10
7 10
7.1 10
7.2 10
7.3 10
7.4 10
7.5 10
7.6 10
7.7 10
7.8 10
7.9 10
8 10
8.1 10
8.2 10
8.3 10
8.4 10
8.5 10
8.6 10
8.7 10
8.8 10
8.9 10
9 10
9.1 10
9.2 10
9.3 10
9.4 10
9.5 10
9.6 10
9.7 10
9.8 10
9.9 10
10 10
10.1 10
10.2 10
10.3 10
10.4 10
10.5 10
10.6 10
10.7 10
10.8 10
10.9 10
11 10
11.1 10
11.2 10
11.3 10
11.4 10
11.5 10
11.6 10
11.7 10
11.8 10
11.9 10
12 10
};
\end{axis}

\begin{axis}[
axis y line=right,
height=\figureheight,
scaled y ticks=false,
tick align=outside,
width=\figurewidth,
x grid style={darkgray176},
xmin=0, xmax=12,
xtick pos=left,
xtick style={color=black},
xticklabel style={align=center},
y grid style={darkgray176},
ylabel=\textcolor{gray}{Distance [\si{\meter}]},
ymin=-11.2, ymax=59.2,
ytick pos=right,
ytick style={color=black},
yticklabel style={/pgf/number format/fixed,/pgf/number format/precision=3},
yticklabel style={anchor=west}
]
\addplot [very thick, gray, dotted]
table {%
0 39.96
0.1 39.56
0.2 39.16
0.3 38.76
0.4 38.36
0.5 37.96
0.6 37.56
0.7 37.16
0.8 36.76
0.9 36.36
1 35.96
1.1 35.56
1.2 35.16
1.3 34.76
1.4 34.36
1.5 33.96
1.6 33.56
1.7 33.16
1.8 32.76
1.9 32.36
2 31.96
2.1 31.56
2.2 31.16
2.3 30.76
2.4 30.36
2.5 29.96
2.6 29.56
2.7 29.16
2.8 28.76
2.9 28.36
3 27.96
3.1 27.5591454618409
3.2 27.1536385714562
3.3 26.7390901729979
3.4 26.3111913636837
3.5 25.8657517399234
3.6 25.3987365916372
3.7 24.9063027146287
3.8 24.3848325231371
3.9 23.8309661597845
4 23.2416313179079
4.1 22.6150201397352
4.2 21.9529378320868
4.3 21.2582214836182
4.4 20.5340251355139
4.5 19.7838048854965
4.6 19.0113010617428
4.7 18.2205175457054
4.8 17.4156983501678
4.9 16.6013015857237
5 15.7819709750126
5.1 14.9625050992113
5.2 14.1478245852362
5.3 13.3429374646077
5.4 12.5529029557532
5.5 11.7827939404547
5.6 11.0376584219859
5.7 10.3224802670573
5.8 9.64213954581617
5.9 9.00137279370816
6 8.40473352586152
6.1 7.85655333871001
6.2 7.36090393475564
6.3 6.92156040462928
6.4 6.54169498967956
6.5 6.22146270789612
6.6 5.95915498257183
6.7 5.75268982853922
6.8 5.5996409726093
6.9 5.49725281235254
7 5.44245740530419
7.1 5.43189338384978
7.2 5.46192667917423
7.3 5.52867292650452
7.4 5.62802141351023
7.5 5.75566042421202
7.6 5.90710382214438
7.7 6.07771870887876
7.8 6.262753987383
7.9 6.4573696541144
8 6.65666663925104
8.1 6.85654793573045
8.2 7.05578260414049
8.3 7.25375882975155
8.4 7.44987255655232
8.5 7.64353121145603
8.6 7.83415735771052
8.7 8.02119225498789
8.8 8.20409930420439
8.9 8.38236735583219
9 8.55551386130625
9.1 8.72308784809665
9.2 8.88467270010417
9.3 9.03988872623718
9.4 9.18839550133334
9.5 9.32989396499405
9.6 9.46412826539122
9.7 9.59088733667883
9.8 9.71000620028409
9.9 9.82136698205523
10 9.92489963899476
10.1 10.0205823910979
10.2 10.1084418556337
10.3 10.1885528830403
10.4 10.2610380954475
10.5 10.3260671306704
10.6 10.3838555963359
10.7 10.43466374059
10.8 10.4787948475813
10.9 10.5165933676133
11 10.5484427934944
11.1 10.5747632961762
11.2 10.5960091342606
11.3 10.6126658533444
11.4 10.6252472924686
11.5 10.6342924161309
11.6 10.6403619913894
11.7 10.6440351305476
11.8 10.6459057207345
11.9 10.6465787623931
12 10.646666639251
};
\end{axis}

\end{tikzpicture}
\begin{tikzpicture}

\definecolor{darkgray176}{RGB}{176,176,176}

\begin{axis}[
height=\figureheight,
scaled y ticks=false,
tick align=outside,
tick pos=left,
width=\figurewidth,
x grid style={darkgray176},
xlabel={Time [\si{\second}]},
xmajorgrids,
xmin=0, xmax=12,
xtick style={color=black},
xticklabel style={align=center},
y grid style={darkgray176},
ylabel={Probability of a crash},
ymajorgrids,
ymin=0, ymax=1,
ytick style={color=black},
yticklabel style={/pgf/number format/fixed,/pgf/number format/precision=3}
]
\addplot [very thick, gray]
table {%
0 4.44089209850063e-16
0.1 4.44089209850063e-16
0.2 4.44089209850063e-16
0.3 7.7715611723761e-16
0.4 9.99200722162641e-16
0.5 1.33226762955019e-15
0.6 1.66533453693773e-15
0.7 2.22044604925031e-15
0.8 3.33066907387547e-15
0.9 4.32986979603811e-15
1 6.10622663543836e-15
1.1 8.21565038222616e-15
1.2 1.11022302462516e-14
1.3 1.50990331349021e-14
1.4 2.05391259555654e-14
1.5 2.80886425230165e-14
1.6 3.85247389544929e-14
1.7 5.26245713672324e-14
1.8 7.24975635080227e-14
1.9 9.96980276113391e-14
2 1.37889699658444e-13
2.1 1.90514271025677e-13
2.2 2.639000129534e-13
2.3 3.66484620428764e-13
2.4 5.10369524420184e-13
2.5 7.123190925995e-13
2.6 9.97313343020778e-13
2.7 1.39921407793508e-12
2.8 1.96831440035794e-12
2.9 2.77644573998259e-12
3 3.92696986040164e-12
3.1 6.40032471466156e-12
3.2 1.37058142613e-11
3.3 3.77178288601954e-11
3.4 1.29081856314883e-10
3.5 5.28062815696728e-10
3.6 2.4757883521076e-09
3.7 1.27664956384166e-08
3.8 6.97261978155339e-08
3.9 3.90337049127609e-07
4 2.1789062505384e-06
4.1 1.10178762287028e-05
4.2 4.72238371914679e-05
4.3 0.00017447317708863
4.4 0.000562841414139981
4.5 0.0016012675891236
4.6 0.00404965713658056
4.7 0.00916556764206033
4.8 0.0186781897985745
4.9 0.0344791495623454
5 0.0580167924716313
5.1 0.0895843620800477
5.2 0.127830768972844
5.3 0.169751724366941
5.4 0.211165886350042
5.5 0.247434745497688
5.6 0.27411892004315
5.7 0.287395820389519
5.8 0.284290204175574
5.9 0.262976760047023
6 0.223521778476522
6.1 0.169293670341529
6.2 0.108544330507501
6.3 0.0540986336254947
6.4 0.0186615586518468
6.5 0.00405153426815508
6.6 0.000436051294375961
6.7 1.49288332450537e-05
6.8 6.4972758728743e-08
6.9 3.3957281431185e-12
7 0
7.1 0
7.2 0
7.3 0
7.4 0
7.5 0
7.6 0
7.7 0
7.8 0
7.9 0
8 0
8.1 0
8.2 0
8.3 0
8.4 0
8.5 0
8.6 0
8.7 0
8.8 0
8.9 0
9 0
9.1 0
9.2 0
9.3 0
9.4 0
9.5 0
9.6 0
9.7 0
9.8 0
9.9 0
10 0
10.1 0
10.2 0
10.3 0
10.4 0
10.5 0
10.6 0
10.7 0
10.8 0
10.9 0
11 0
11.1 0
11.2 0
11.3 0
11.4 0
11.5 0
11.6 0
11.7 0
11.8 0
11.9 0
12 0
};
\addplot [very thick, black]
table {%
0 5.98598443864919e-05
0.1 5.97315200644449e-05
0.2 5.96685792819155e-05
0.3 5.96774730194659e-05
0.4 5.97654703311494e-05
0.5 5.99407296333559e-05
0.6 6.02123736923254e-05
0.7 6.05905680130037e-05
0.8 6.10866022024672e-05
0.9 6.17129737323837e-05
1 6.24834733570863e-05
1.1 6.34132712580455e-05
1.2 6.45190027844299e-05
1.3 6.58188524474615e-05
1.4 6.73326346100528e-05
1.5 6.90818691021935e-05
1.6 7.10898497995121e-05
1.7 7.33817040440434e-05
1.8 7.59844406836249e-05
1.9 7.89269844856417e-05
2 8.22401947735142e-05
2.1 8.59568663774425e-05
2.2 9.01117114271205e-05
2.3 9.47413211912897e-05
2.4 9.98841081394991e-05
2.5 0.000105580229721025
2.6 0.000111871497082112
2.7 0.000118801274132747
2.8 0.000126414375082788
2.9 0.000134756971843893
3 0.000143876526580703
3.1 0.000439560036242466
3.2 0.000995921371000775
3.3 0.00157891094518547
3.4 0.0022678817381056
3.5 0.00431542019650416
3.6 0.0116901976915259
3.7 0.0299357132272735
3.8 0.0633893693543778
3.9 0.119102614913533
4 0.198813781827371
4.1 0.294440488604766
4.2 0.393394545586453
4.3 0.484281629215184
4.4 0.562189997112093
4.5 0.628235198033254
4.6 0.686032304120968
4.7 0.738946707785941
4.8 0.788595637081177
4.9 0.834112493762541
5 0.871809066235264
5.1 0.895220218306859
5.2 0.898499629376688
5.3 0.883836091097371
5.4 0.855670260359818
5.5 0.816397678771563
5.6 0.785027580302037
5.7 0.735370943224077
5.8 0.580714280134561
5.9 0.407749942000074
6 0.354781345040706
6.1 0.359011844080659
6.2 0.346365618216123
6.3 0.298989842821921
6.4 0.252527712290496
6.5 0.217484576002841
6.6 0.183702659297321
6.7 0.152441898645763
6.8 0.124421106534954
6.9 0.100005199699694
7 0.0793159492454853
7.1 0.062267372549629
7.2 0.0485891704471518
7.3 0.0378751251689326
7.4 0.0296514551690476
7.5 0.0234430261743263
7.6 0.0188206667560959
7.7 0.0154251976166538
7.8 0.0129728049135095
7.9 0.0112495101724166
8 0.0101015272380661
8.1 0.00927209829539462
8.2 0.00856986148054158
8.3 0.00797942772768136
8.4 0.00748690055931797
8.5 0.00707993348314635
8.6 0.00674771229797538
8.7 0.0064808798781107
8.8 0.00627142215459702
8.9 0.00611253226685204
9 0.00599846661096386
9.1 0.00592440277980979
9.2 0.00588630582447028
9.3 0.00588080622071844
9.4 0.00590509054756056
9.5 0.00595680419650023
9.6 0.0060339643765075
9.7 0.00613488116987342
9.8 0.0062580843231814
9.9 0.00640225371931374
10 0.00656615197080137
10.1 0.00674855821239377
10.2 0.00694820287444916
10.3 0.00716370392479004
10.4 0.00739350572399629
10.5 0.00763582220814856
10.6 0.00788858656374931
10.7 0.00814940986918834
10.8 0.00841555132819367
10.9 0.00868390269960488
11 0.00895098932438099
11.1 0.00921298975907149
11.2 0.00946577544502598
11.3 0.00970497108269778
11.4 0.00992603545965923
11.5 0.010124361431543
11.6 0.0102953926234065
11.7 0.0104347532654504
11.8 0.0105383864744286
11.9 0.0106026953222264
12 0.010624680281447
};
\end{axis}

\end{tikzpicture}\\
\begin{tikzpicture}

\definecolor{darkgray176}{RGB}{176,176,176}

\begin{axis}[
height=\figureheight,
scaled y ticks=false,
tick align=outside,
tick pos=left,
width=\figurewidth,
x grid style={darkgray176},
xlabel={Time [\si{\second}]},
xmajorgrids,
xmin=0, xmax=6,
xtick style={color=black},
xticklabel style={align=center},
y grid style={darkgray176},
ylabel=\textcolor{black}{Speed [\si{\meter\per\second}]},
ymajorgrids,
ymin=9.55695328945044, ymax=24.6877641290738,
ytick style={color=black},
yticklabel style={/pgf/number format/fixed,/pgf/number format/precision=3}
]
\addplot [very thick, black]
table {%
0 24
0.1 24
0.2 24
0.3 24
0.4 24
0.5 24
0.6 24
0.7 24
0.8 24
0.9 24
1 24
1.1 24
1.2 24
1.3 24
1.4 24
1.5 24
1.6 24
1.7 24
1.8 24
1.9 24
2 24
2.1 24
2.2 24
2.3 24
2.4 24
2.5 24
2.6 24
2.7 24
2.8 24
2.9 24
3 24
3.1 24
3.2 24
3.3 24
3.4 24
3.5 24
3.6 24
3.7 24
3.8 24
3.9 24
4 24
4.1 23.975338669865
4.2 23.9015067247611
4.3 23.7789593631814
4.4 23.6084521303612
4.5 23.3910362600903
4.6 23.1280521935069
4.7 22.8211213148327
4.8 22.4721359549996
4.9 22.0832477248002
5 21.6568542494924
5.1 21.1955843866415
5.2 20.7022820183398
5.3 20.1799885177276
5.4 19.6319239979164
5.5 19.0614674589207
5.6 18.4721359549996
5.7 17.8675629108472
5.8 17.2514757203218
5.9 16.6276727658228
6 16
};
\addplot [very thick, black, dashed]
table {%
0 20
0.1 20
0.2 20
0.3 20
0.4 20
0.5 20
0.6 20
0.7 20
0.8 20
0.9 20
1 20
1.1 20
1.2 20
1.3 20
1.4 20
1.5 20
1.6 20
1.7 20
1.8 20
1.9 20
2 20
2.1 20
2.2 20
2.3 20
2.4 20
2.5 20
2.6 20
2.7 20
2.8 20
2.9 20
3 20
3.1 19.9778098230154
3.2 19.9114362536434
3.3 19.8014684283847
3.4 19.6488824294413
3.5 19.4550326209418
3.6 19.2216396275101
3.7 18.9507750618785
3.8 18.6448431371071
3.9 18.3065593266183
4 17.9389262614624
4.1 17.5452070787519
4.2 17.1288964578254
4.3 16.6936896012265
4.4 16.2434494358243
4.5 15.7821723252012
4.6 15.3139525976466
4.7 14.8429462046094
4.8 14.3733338321785
4.9 13.9092837930173
5 13.4549150281253
5.1 13.0142605468261
5.2 12.5912316294914
5.3 12.1895831107393
5.4 11.8128800512565
5.5 11.4644660940673
5.6 11.1474337861211
5.7 10.8645971286272
5.8 10.6184665997807
5.9 10.4112268715801
6 10.2447174185242
};
\end{axis}

\begin{axis}[
axis y line=right,
height=\figureheight,
scaled y ticks=false,
tick align=outside,
width=\figurewidth,
x grid style={darkgray176},
xmin=0, xmax=6,
xtick pos=left,
xtick style={color=black},
xticklabel style={align=center},
y grid style={darkgray176},
ylabel=\textcolor{gray}{Distance [\si{\meter}]},
ymin=-1.77218684219822, ymax=58.7510565162952,
ytick pos=right,
ytick style={color=black},
yticklabel style={/pgf/number format/fixed,/pgf/number format/precision=3},
yticklabel style={anchor=west}
]
\addplot [very thick, gray, dotted]
table {%
0 31.5553
0.1 31.1553
0.2 30.7553
0.3 30.3553
0.4 29.9553
0.5 29.5553
0.6 29.1553
0.7 28.7553
0.8 28.3553
0.9 27.9553
1 27.5553
1.1 27.1553
1.2 26.7553
1.3 26.3553
1.4 25.9553
1.5 25.5553
1.6 25.1553
1.7 24.7553
1.8 24.3553
1.9 23.9553
2 23.5553
2.1 23.1553
2.2 22.7553
2.3 22.3553
2.4 21.9553
2.5 21.5553
2.6 21.1553
2.7 20.7553
2.8 20.3553
2.9 19.9553
3 19.5553
3.1 19.1544454618409
3.2 18.7489385714562
3.3 18.3343901729979
3.4 17.9064913636837
3.5 17.4610517399234
3.6 16.9940365916372
3.7 16.5016027146287
3.8 15.9801325231372
3.9 15.4262661597845
4 14.8369313179079
4.1 14.2103201397352
4.2 13.5482378320868
4.3 12.8535214836182
4.4 12.1293251355139
4.5 11.3791048854965
4.6 10.6066010617428
4.7 9.81581754570538
4.8 9.01099835016781
4.9 8.19660158572374
5 7.37727097501256
5.1 6.55780509921125
5.2 5.74312458523616
5.3 4.93823746460766
5.4 4.14820295575325
5.5 3.37809394045469
5.6 2.63295842198592
5.7 1.91778026705734
5.8 1.23743954581617
5.9 0.596672793708166
6 3.35258615180578e-05
};
\end{axis}

\end{tikzpicture}
\begin{tikzpicture}

\definecolor{darkgray176}{RGB}{176,176,176}

\begin{axis}[
height=\figureheight,
scaled y ticks=false,
tick align=outside,
tick pos=left,
width=\figurewidth,
x grid style={darkgray176},
xlabel={Time [\si{\second}]},
xmajorgrids,
xmin=0, xmax=6,
xtick style={color=black},
xticklabel style={align=center},
y grid style={darkgray176},
ylabel={Probability of a crash},
ymajorgrids,
ymin=0, ymax=1,
ytick style={color=black},
yticklabel style={/pgf/number format/fixed,/pgf/number format/precision=3}
]
\addplot [very thick, gray]
table {%
0 1.91069382537989e-13
0.1 2.64899213675562e-13
0.2 3.68038932663239e-13
0.3 5.12145881259585e-13
0.4 7.15094650161063e-13
0.5 1.00119912360697e-12
0.6 1.40498723766314e-12
0.7 1.9764190284377e-12
0.8 2.78754797022884e-12
0.9 3.94284604965378e-12
1 5.5916382635246e-12
1.1 7.95230548078507e-12
1.2 1.13412612634534e-11
1.3 1.62205804343785e-11
1.4 2.32677210831866e-11
1.5 3.34750005492879e-11
1.6 4.83060258460455e-11
1.7 6.99217350685899e-11
1.8 1.01527009022107e-10
1.9 1.47887369017496e-10
2 2.16114348638996e-10
2.1 3.16857984294927e-10
2.2 4.66120808617632e-10
2.3 6.88034296203455e-10
2.4 1.01911934446974e-09
2.5 1.51484813581959e-09
2.6 2.2597909099531e-09
2.7 3.38336358929325e-09
2.8 5.08436626134312e-09
2.9 7.66936092411186e-09
3 1.16129417193633e-08
3.1 1.98656593397573e-08
3.2 4.2851094117502e-08
3.3 1.14110764615205e-07
3.4 3.63827501348446e-07
3.5 1.33897443010955e-06
3.6 5.47009243823116e-06
3.7 2.3865944629109e-05
3.8 0.00010722474571323
3.9 0.00047963486539726
4 0.0020696478379757
4.1 0.00792597426473152
4.2 0.025463829112441
4.3 0.0689977150121731
4.4 0.157929108698798
4.5 0.305524766244427
4.6 0.50111630703065
4.7 0.703568875521019
4.8 0.862550296312246
4.9 0.953575935152977
5 0.989543345439849
5.1 0.998612169658351
5.2 0.999910346901267
5.3 0.99999797664004
5.4 0.999999991894758
5.5 0.999999999999002
5.6 1
5.7 1
5.8 1
5.9 1
6 1
};
\addplot [very thick, black]
table {%
0 8.60030719474686e-05
0.1 9.01632794047851e-05
0.2 9.47986924816148e-05
0.3 9.99477446269444e-05
0.4 0.000105650613276093
0.5 0.000111949128418362
0.6 0.000118886671592839
0.7 0.000126508073467275
0.8 0.000134859521438557
0.9 0.000143988492587637
1 0.000153943731859689
1.1 0.000164775300544745
1.2 0.000176534726018912
1.3 0.0001892752902795
1.4 0.000203052502076273
1.5 0.000217924805421936
1.6 0.0002339545859943
1.7 0.000251209546493198
1.8 0.000269764532513401
1.9 0.000289703902140685
2 0.000311124545566104
2.1 0.00033413967595239
2.2 0.000358883530119949
2.3 0.000385517138046843
2.4 0.000414235344575271
2.5 0.000445275296178428
2.6 0.000478926641532756
2.7 0.00051554373870999
2.8 0.000555560216374502
2.9 0.000599506304673634
3 0.00064802943822761
3.1 0.00219777423969306
3.2 0.00565766319738377
3.3 0.0110191131423752
3.4 0.0198655697232279
3.5 0.0387848318002267
3.6 0.088256358020174
3.7 0.173450994821469
3.8 0.273509328174698
3.9 0.383003548483483
4 0.499167876354031
4.1 0.617012621254554
4.2 0.7269302156634
4.3 0.817579516643046
4.4 0.88418851657621
4.5 0.929361342301937
4.6 0.958721022948261
4.7 0.977355611490616
4.8 0.988715066827874
4.9 0.995051469792295
5 0.998072450229731
5.1 0.999178250741368
5.2 0.999503206464657
5.3 0.999724530839781
5.4 0.999906402452967
5.5 0.999933379652549
5.6 0.999757104301946
5.7 0.999431851605672
5.8 0.998026337387207
5.9 0.989803303704762
6 1
};
\end{axis}

\end{tikzpicture}
	\caption{Demonstration of \acp{ssm} for 3 hypothetical scenarios. 
		The left plots show the speeds of the ego vehicle (solid black line) and leading vehicle (dashed black line) and the distance between the ego vehicle and the leading vehicle (dotted gray line, scale on the right of the plot).
		The right plots show the estimated probability of a crash corresponding to the 3 scenarios according to the \ac{ssm} explained in \cref{sec:ngsim metric} (black lines) and the \ac{ssm} of \textcite{wang2014evaluation} explained in \cref{sec:wang stamatiadis explanation} (gray lines).}
	\label{fig:scenarios}		
\end{figure}

The first scenario in \cref{fig:scenarios} (top row) shows a scenario in which the leading vehicle initially drives with a speed of \SI{20}{\meter\per\second}.
The leading vehicle starts to decelerate after \SI{3}{\second} toward a speed of \SI{10}{\meter\per\second} with an average deceleration of \SI{3}{\meter\per\second\squared}.
The ego vehicle initially drives with a speed of \SI{24}{\meter\per\second} at a distance of \SI{40}{\meter} from the leading vehicle.
The ego vehicle starts decelerating after \SI{2}{\second} toward a speed of \SI{8}{\meter\per\second} within \SI{4}{\second}.
It takes \SI{4}{\second} more to reach the speed of the leading vehicle.
Because the ego vehicle always maintains a relatively large distance toward the leading vehicle, both \acp{ssm} do not qualify this scenario as risky, considering the estimated crash probability that stays below 0.1.

The second scenario in \cref{fig:scenarios} (center row) differs from the first scenario in that the ego vehicle starts to decelerate \SI{2}{\second} later.
As a result, the ego vehicle approaches the leading vehicle up to a distance of \SI{5.4}{\meter}.
According to $\probabilityestcond{\collision}{\situationinitial}$ from \cref{sec:ngsim metric} (black line in the right plot of \cref{fig:scenarios}), the probability of a crash reaches almost 1, indicating that around that time, the risk of a crash is high.
The local minimum of $\probabilityestcond{\collision}{\situationinitial}$ at around \SI{6}{\second} illustrates the effect of the numerical approximation of $\probabilitycond{\collision}{\situationinitial}$.
Because we have used $\simulationthreshold=0.1>0$, the resulting estimation may have an error. 
When lowering the threshold $\simulationthreshold$, the resulting $\probabilityestcond{\collision}{\situationinitial}$ in the center right plot in \cref{fig:scenarios} will be smoother. 
This goes, however, at the cost of an increased number of simulations\footnote{Alternatively, the bandwidth matrix $\bandwidthnw$ may be increased. 
	On the one hand, this will lower the variance of the error, but, on the other hand, it will increase the bias of the result.
	We refer the interested reader to \autocite{chen2017tutorial} for more details on the effect of $\bandwidthnw$.}.

The third scenario in \cref{fig:scenarios} (bottom row) differs from the second scenario in that the initial distance between the ego vehicle and the leading vehicle is \SI{31.5}{\meter} instead of \SI{40}{\meter}. 
As a result, the ego vehicle collides with the leading vehicle after \SI{6}{\second}.
As expected, the \acp{ssm} in \cref{fig:scenarios} indicate a crash probability of 1.
The difference between $\probabilityestcond{\collision}{\situationinitial}$ and $\wangstamatiadis$ is that $\probabilityestcond{\collision}{\situationinitial}$ increases earlier. 
Note that $\probabilityestcond{\collision}{\situationinitial}$ increasing sooner than $\wangstamatiadis$ does not necessarily mean that it is better: because there is no objective truth for \iac{ssm}, we cannot argue that one \ac{ssm} is better than another \ac{ssm}.
Hence, in the next section, we will present a quantitative approach to benchmark \iac{ssm}.

\cstart Note that the choice of $\dimension=4$ can be considered to be somewhat arbitrary. 
Therefore, we have repeated this case study described with $\dimension=3$ and $\dimension=5$. 
With $\dimension=3$, the results are comparable to the results with $\dimension=4$ and there is hardly any difference noticeable.
With $\dimension=5$, the derived \ac{ssm} fluctuates a bit more, which resulted in a less smooth \ac{ssm}.
Most likely, this is the case because the \ac{kde} becomes less reliable for $\dimension=5$.
We cannot argue objectively which choice of $\dimension$ is better, but since a lower $\dimension$ generally leads to more loss of information and the \ac{ssm} with $\dimension=5$ is less smooth, we have opted for $\dimension=4$. \cend

\subsection{Benchmarking an SSM with expected risk trends}
\label{sec:trends}

In this section, we demonstrate an approach for benchmarking \iac{ssm} that is based on expected risk trends discussed in \textcite{mullakkal2017comparative}, who argue that the risk increases if the approaching speed of the ego vehicle toward the leading vehicle increases.
\cstart Furthermore, the risk of the ego vehicle colliding with its leading vehicle increases with a higher ego vehicle speed \autocite{aarts2006driving}.
Note that as a result of the first risk trend, the risk of a collision with a vehicle behind the ego vehicle decreases with high ego vehicle speed, but since the current benchmarking only considers the risk of the ego vehicle colliding with its leading vehicle, the risk of a collision with a vehicle behind the ego vehicle is not further considered.
The risk also increases with a \cend higher driver reaction time \autocite{klauer2006impact}.
On the other hand, the risk decreases with a higher road friction \autocite{wallman2001friction} or a larger intervehicle spacing \autocite{mullakkal2017comparative}.

To check whether the developed \ac{ssm} follows these 5 expected risk trends\footnote{In \autocite{mullakkal2017comparative}, a sixth expected risk trend is mentioned based on \autocite{evans1994driver}, namely the vehicle mass. 
	Our interpretation of \autocite{evans1994driver}, however, is that the ratio of masses of two colliding vehicles influences the safety risk and that one cannot argue that a higher mass of the ego vehicle necessarily increases the safety risk. 
	Therefore, we exclude the ego vehicle mass from our analysis.}, we evaluate the partial derivatives of the measure of \cref{eq:nadaraya watson}.
The intuition is as follows: If the expected risk trend for an input X (e.g., the ego vehicle speed) is that the risk increases as X increases, then we expect the partial derivative of our \ac{ssm} with respect to X to be positive.
Furthermore, if we evaluate the partial derivative at many points, we expect that at least the majority of these evaluated partial derivatives is positive.
Similarly, if we expect that the risk measure decreases with increasing X, then we expect that at least the majority of the evaluated partial derivatives is negative.
\cstart Note that because the proposed benchmarking method only considers the partial derivatives, no claim can be made regarding the actual accuracy of the \ac{ssm}. \cend

To illustrate the approach for benchmarking \iac{ssm}, we use the \ac{ssm} of \cref{sec:ngsim metric} with a few different assumptions.
Because we have not described an expected trend regarding $\accelerationleadsymbol$, we simply use $\accelerationleadsymbol=0$.
Also, because the expected risk trend for the relative speed is defined, we use the relative speed, i.e., $\speeddifferencesymbol=\speedegosymbol-\speedleadsymbol$, instead of $\speedleadsymbol$. 
For the same reason, instead of assuming a random reaction time $\timereact$ and \ac{madr} $\accelerationmax$, these are now considered as input to our measure. 
Finally, instead of using the log of the gap between the ego vehicle and the leading vehicle, we use the gap as a direct input.
Thus, we have:
\begin{equation}
	\label{eq:input partial derivatives}
	\situationinitial\transpose = \begin{bmatrix}
		\speedegosymbol-\speedleadsymbol & \speedegosymbol & \timereact & \gapsymbol & \accelerationmax
	\end{bmatrix}.
\end{equation}

We compute $\probabilityestcond{\collision}{\situationinitial}$ using \cref{eq:nadaraya watson} where the points $\left\{\situationinitialinstance{\situationindexdesign}'\right\}_{\situationindexdesign=1}^{\numberofdesignpoints}$ are taken from a grid.
For each input variable, 10 different values at equal distance are used, resulting in $\numberofdesignpoints=10^5$.
Here, $\speedegosymbol-\speedleadsymbol$ ranges from \SI{0}{\meter\per\second} to \SI{20}{\meter\per\second}, $\speedegosymbol$ ranges from \SI{10}{\meter\per\second} to \SI{30}{\meter\per\second}, $\timereact$ ranges from \SI{0.5}{\second} to \SI{1.5}{\second}, $\gapsymbol$ ranges from \SI{5}{\meter} to \SI{30}{\meter}, and $\accelerationmax$ ranges from \SI{4}{\meter\per\second\squared} to \SI{10}{\meter\per\second\squared}.
A threshold $\simulationthreshold=0.02$ is used.
For the bandwidth matrix $\bandwidthnw$, we use a diagonal matrix with the $(i,i)$-th entry corresponding to the squared difference between two consecutive values of the $i$-th entry of $\situationinitial$.
For example, the first value is $(\SI{20}{\meter\per\second}/(10-1))^2 \approx \SI{4.9}{\meter\squared\per\second\squared}$. 
The other values on the diagonal are: \SI{4.9}{\meter\squared\per\second\squared}, \SI{0.012}{\second\squared}, \SI{7.7}{\meter\squared}, and \SI{0.44}{\meter\squared\per\second\tothe{4}}.
For each input variable listed in \cref{eq:input partial derivatives}, we evaluate the partial derivative of \cref{eq:nadaraya watson} at each  $\situationinitialinstance{\situationindexdesign}'$, $\situationindexdesign\in\{1,\ldots,\numberofdesignpoints\}$.

\Cref{tab:trends} shows the result of the benchmarking. 
It shows that the \ac{ssm} follows the expected risk trends mostly. 
E.g., in more than \SI{99}{\%} of the cases, the partial derivative of the relative speed ($\speedegosymbol-\speedleadsymbol$) is positive.
For the remaining \SI{1}{\%}, the partial derivative is negative, albeit only slightly. 
One explanation is that this remaining \SI{1}{\percent} is caused by the inaccuracies introduced by the numerical approximation of \cref{eq:estimate probability of collision}.

\begin{table}
	\centering
	\caption{Percentiles of the partial derivatives of the \ac{ssm} and the corresponding expected risk trends.}
	\label{tab:trends}
	\begin{tabular}{lrrrrr}
		\toprule
		& $\speedegosymbol-\speedleadsymbol$ & $\speedegosymbol$ & $\timereact$ & $\gapsymbol$ & $\accelerationmax$ \\
		\otoprule
		Expected trend & Increase & Increase & Increase & Decrease & Decrease \\
		Maximum         &  0.1629 &  0.1555 &  1.3136 &  0.0010 &  0.0037 \\
		99th percentile &  0.1585 &  0.1162 &  0.8968 &  0.0002 &  0.0002 \\
		95th percentile &  0.1495 &  0.0524 &  0.6765 &  0.0000 & -0.0000 \\
		90th percentile &  0.1346 &  0.0151 &  0.5351 & -0.0000 & -0.0000 \\
		75th percentile &  0.0746 &  0.0012 &  0.2917 & -0.0002 & -0.0003 \\
		50th percentile &  0.0114 &  0.0001 &  0.0605 & -0.0070 & -0.0054 \\
		25th percentile &  0.0004 &  0.0000 &  0.0022 & -0.0320 & -0.0290 \\
		10th percentile &  0.0000 & -0.0000 &  0.0000 & -0.0545 & -0.0654 \\
		 5th percentile &  0.0000 & -0.0002 &  0.0000 & -0.0645 & -0.0880 \\
		 1st percentile &  0.0000 & -0.0007 & -0.0020 & -0.0781 & -0.1337 \\
		Minimum         & -0.0035 & -0.0030 & -0.0180 & -0.1076 & -0.2030 \\
		\bottomrule
	\end{tabular}
\end{table}

\section{Discussion}
\label{sec:discussion}

Typically, \acp{ssm} rely on assumptions regarding the behavior of traffic participants. 
An advantage of the presented \ac{ourmethod} method for deriving \acp{ssm} is that the \ac{ourmethod} method is not bound to certain predetermined assumptions. 
We want to stress, however, that when using the \ac{ourmethod} method for deriving \iac{ssm}, a set of assumptions is still needed.
In fact, multiple \acp{ssm} can be derived by using the \ac{ourmethod} method with different sets of assumptions. 
As a result, the \ac{ourmethod} method can be used to derive multiple \acp{ssm} that are applicable in various types of scenarios, e.g., ranging from vehicle-following scenarios to scenarios at intersections. 
Note that although the \ac{ourmethod} method is applicable in various types of scenarios, the current case study focuses on longitudinal traffic conflicts.
In a future work, we will present the application of the \ac{ourmethod} method for deriving \acp{ssm} for lateral traffic conflicts.

The \ac{ourmethod} method uses data to adapt the \acp{ssm} to, e.g., the local traffic behavior. 
More specifically, the data are used to predict the possible future situations ($\situationfuture$) given an initial situation ($\situationinitial$).
This can be an advantage because the data can be used to rely less on assumptions as to how the future develops given an initial situation. 
To fully benefit from this approach, the data should satisfy a few conditions.
First, the recorded data need to represent the actual traffic behavior in which the \acp{ssm} are applied. 
Second, we need enough data to estimate $\densitycond{\situationfuture}{\situationinitial}$.
In \autocite{degelder2019completeness}, a metric is presented that can be used to determine whether enough data have been collected to estimate $\densitycond{\situationfuture}{\situationinitial}$ accurately.

The \ac{ourmethod} method can still be applied in case no data are available.
The first alternative is to use existing knowledge to determine an estimate of $\densitycond{\situationfuture}{\situationinitial}$ instead of estimating $\densitycond{\situationfuture}{\situationinitial}$ on the basis of data. 
For example, statistics or literature on driving behavior of traffic participants may be used.
The second alternative is to use assumptions on how the future develops given an initial situation $\situationinitial$.
For example, when assuming that the speed of the leading vehicle in \cref{sec:wang stamatiadis replicate} remains constant, it is not needed to estimate $\densitycond{\situationfuture}{\situationinitial}$.
Note that a combination is also possible.
For example, estimate $\densitycond{\situationfuture}{\situationinitial}$ based on data in case $\situationinitial$ is well represented in the data, but define $\densitycond{\situationfuture}{\situationinitial}$ on the basis of existing knowledge and/or assumptions for the cases where $\situationinitial$ is underrepresented in the data.
\cstart A third alternative is to use other methods for predicting the trajectories of the other traffic participants, e.g., using hidden Markov models \autocite{laugier2011probabilistic}, sequence similarity methods \autocite{saunier2007probabilistic}, Gaussian mixture models \autocite{wiest2012probabilistic}, or long short-term memory networks \autocite{deo2018multi}. 
For an overview of trajectory prediction models for vehicles and pedestrians, see \autocite{lefevre2014survey} and \autocite{rudenko2020human}, respectively. \cend

Note that the \ac{ourmethod} method is used to derive \acp{ssm} that predict the probability of a specific event, such as a crash, i.e., the derived \acp{ssm} can be used as a measure of proximity of the specified event.
However, the \ac{ourmethod} method is not used to measure the severity of an interaction, i.e., the extent of harm in case the interaction leads to a crash.
For measuring the severity of an interaction, typically energy-based \acp{ssm} are used \autocite{wang2021review}.
So, if there is a need to also have an indicator of the severity of an interaction, an energy-based \ac{ssm}, e.g., see \autocite{ozbay2008derivation, alhajyaseen2015integration, laureshyn2017search, mullakkal2020probabilistic}, may be considered alongside \iac{ssm} derived using the \ac{ourmethod} method.

We have illustrated the \ac{ourmethod} method through different derived \acp{ssm} in the case study.
The derived \acp{ssm} estimate the probability of a crash with a leading vehicle under different assumptions.
Because of the focus on crashes, the resulting \acp{ssm} may still be low given an initial situation that is generally considered to be unsafe. 
For example, the \ac{ssm} described in \cref{sec:ngsim metric} gives a crash probability of approximately \SI{14}{\percent} when approaching a leading vehicle that is driving at a constant speed of $\speedleadsymbol=\SI{12}{\meter\per\second}$ ($\accelerationleadsymbol=\SI{0}{\meter\per\second\squared}$) with a speed of $\speedegosymbol=\SI{25}{\meter\per\second}$ and a gap of $\gapsymbol=\SI{20}{\meter}$ (see left heat map in \cref{fig:heatmaps}).
In this initial situation, the \ac{thw} is only $\gapsymbol/\speedegosymbol=\SI{0.8}{\second}$ and the \ac{ttc} is only $\gapsymbol/(\speedegosymbol-\speedleadsymbol)=\SI{1.5}{\second}$, whereas \iac{thw} of less than \SI{1}{\second} or \iac{ttc} of less than \SI{1.5}{\second} is considered unsafe \autocite{vogel2003comparison}.
In order to put more emphasis on such unsafe situations, different events --- instead of crashes --- can be considered.
For example, we can derive \iac{ssm} that estimates the probability that the \ac{ttc} is below \SI{1}{\second} within the next 5 seconds.
More research is needed to investigate whether such \acp{ssm} can be of practical use, e.g., for evaluating whether a driver is actively pursuing large safety margins.

A few choices have to be made when using the \ac{ourmethod} method for deriving \acp{ssm}.
One such a choice is the set of initial situations $\{\situationinitialinstance{1},\ldots,\situationinitialinstance{\numberofdesignpoints}\}$ for which the probability $\probabilitycond{\collision}{\situationinitial}$ is estimated.
Generally speaking, for larger $\numberofdesignpoints$, the approximation of $\probabilitycond{\collision}{\situationinitial}$ in \cref{eq:nadaraya watson} improves.
One disadvantage, however, is that more simulation runs are required when $\numberofdesignpoints$ is larger, but because these simulation runs are performed offline, this problem might be solved by, e.g., parallel computing resources. 
Another disadvantage is that the computational cost of the approximation in \cref{eq:nadaraya watson} scales linearly with $\numberofdesignpoints$.
Especially when using this approximation for real-time evaluation of the \ac{ssm}, this can be a bottleneck.
One solution to this is to not use all $\numberofdesignpoints$ initial situations for evaluating \cref{eq:nadaraya watson}.
The intuition is as follows: since \cref{eq:nadaraya watson} uses local regression, an initial situation $\situationinitialinstance{\situationindexdesign}$ can be removed from the set $\{\situationinitialinstance{1},\ldots,\situationinitialinstance{\numberofdesignpoints}\}$ if all neighboring data points give (approximately) the same probability of the event $\collision$, i.e., $|\probabilityestcond{\collision}{\situationinitialinstance{\situationindex}}-\probabilityestcond{\collision}{\situationinitialinstance{\situationindexdesign}}|$ is below a threshold for all $\situationinitialinstance{\situationindex}$, $\situationindex\ne\situationindexdesign$ for which $\normtwo{\situationinitialinstance{\situationindex}-\situationinitialinstance{\situationindexdesign}}$ is below another threshold (assuming that $\probabilityestcond{\collision}{\situationinitial}$ is sufficiently smooth).
For example, the \ac{ssm} that is shown in \cref{fig:heatmaps}, only a few initial situations are required in the upper right region of the heat maps, since the estimated probability is always lower than 0.1.

Another choice is the threshold $\simulationthreshold$ that controls the number of simulation runs ($\numberofsimulations$) that are used to estimate $\probabilitycond{\collision}{\situationinitial}$.
According to \cref{eq:condition stop simulations}, $\numberofsimulations$ is increased until the variance of the estimation error is below $\simulationthreshold$, i.e., $\variance{\probabilitycond{\collision}{\situationinitial}-\probabilityestcond{\collision}{\situationinitial}}<\simulationthreshold$.
Therefore, a lower $\simulationthreshold$ generally results in more accurate estimations of the probability, as illustrated in \cref{fig:ws comparison}.
The downside, however, is that for a lower $\simulationthreshold$, more offline simulation runs are required. 
Although a good choice of $\simulationthreshold$ remains a topic of research, based on experience, we advice to use a maximum threshold of $\simulationthreshold=0.1$ and lower values if the computational resources allow for this.

In the examples presented in \cref{sec:case study}, we have considered the leading vehicle as the only traffic participant other than the ego vehicle.
The \ac{ourmethod} method can be applied in scenarios with multiple traffic participants other than the ego vehicle.
However, the number of parameters ($\situationinitialdim$, i.e., the size of $\situationinitial$) then becomes larger.
As a result, two problems may arise.
First, as $\numberofdesignpoints$ grows exponentially with $\situationinitialdim$, so does the number of simulation runs.
Second, even if these simulation runs can be performed, the regression using \cref{eq:nadaraya watson} becomes slow due to the large $\numberofdesignpoints$.
To overcome these problems, \iac{ssm} can be computed for each traffic participant independently.
For example, let $\numberoftrafficparticipants$ denote the number of traffic participants other than the ego vehicle.
With $\trafficparticipantindex\in\{1,\ldots,\numberoftrafficparticipants\}$, let $\collision_{\trafficparticipantindex}$ denote the event of colliding with the $\trafficparticipantindex$-th traffic participant and let $\situationinitialtp{\trafficparticipantindex}$ denote the initial situation considering the $\trafficparticipantindex$-th traffic participant.
Under the assumption that $\probabilitycond{\collision_{\trafficparticipantindex}}{\situationinitialtp{\trafficparticipantindex}}$ is independent of $\situationinitialtp{\trafficparticipantindexb}$ for all $\trafficparticipantindex\ne\trafficparticipantindexb$, we can calculate the probability of colliding with one or more traffic participants using
\begin{equation}
	\label{eq:combine traffic participants}
	1 - \prod_{\trafficparticipantindex=1}^{\numberoftrafficparticipants}
	\left( 1 - \probabilitycond{\collision_{\trafficparticipantindexb}}{\situationinitialtp{\trafficparticipantindexb}} \right).
\end{equation}
For example, consider a scenario with multiple crossing pedestrians. 
Using the \ac{ourmethod} method, we can derive \iac{ssm} that estimates the probability of colliding with a pedestrian.
Then, after evaluating this \ac{ssm} for each pedestrian, the probability of colliding with one or more pedestrians can be calculated using \cref{eq:combine traffic participants} without the need for \iac{ssm} that considers multiple pedestrians.

In the case study, we have shown how to analyze \iac{ssm} both qualitatively, using heat maps and testing the \ac{ssm} in different scenarios, and quantitatively by benchmarking the \ac{ssm} with expected risk trends \autocite{mullakkal2017comparative}.
Since the \acp{ssm} derived using the \ac{ourmethod} method provide a probability, it is also possible to verify the estimated probability by comparing it with real data. 
This requires, however, an extensive data set that would allow for estimating the probability of the event $\collision$, e.g., a crash, in the near future given a certain situation a vehicle is in. 
It remains a topic for future work to use such a data set to verify the \acp{ssm} derived using the \ac{ourmethod} method.

\cstart The \ac{ourmethod} method is a novel approach for deriving probabilistic \acp{ssm} for risk evaluation. 
Some limitations, however, may hamper its use for real-world applications.
First, as described earlier, if the set of initial situations $\{\situationinitialinstance{1},\ldots,\situationinitialinstance{\numberofdesignpoints}\}$ is too large, real-time calculation of the \ac{ssm} may be difficult. 
As a consequence, the dimension of the vector describing the initial situation, $\situationinitial$, cannot be too large, meaning that the initial situation needs to be encoded into a limited set of numbers.
Second, \ac{kde} does not work well for large dimensions.
Note, however, that we have provided some options for reducing the dimensionality, and, if reducing the dimensionality further is not an good option, the \ac{ourmethod} method can also be applied when other methods are used for the probability density estimation.
Third, many simulations may be required. 
It helps that the simulations can be conducted offline, but it may still be challenging to conduct the simulations in a reasonable time window.
Fourth, although we claim that the \ac{ourmethod} method can be used to derive multiple \acp{ssm} for different type of scenarios, we cannot claim that the derived \acp{ssm} are more valid than others \acp{ssm}, nor can we derive \iac{ssm} that is valid for all types of scenarios.
Lastly, for the simulations, the response of the ego vehicle must be assumed.
This may not coincide with the ego vehicle response in reality.
In a future work, in case of a human driver, this may be tackled by considering the state of the human driver, such as whether the eyes are on the road and/or towards a conflicting traffic participant, as part of the state vector that is used to describe the initial situation. \cend

\acresetall
\section{Conclusions}
\label{sec:conclusions}

Road safety is an important research topic.
To quantify the safety at a vehicle level, \acp{ssm} are often used to characterize the risk of a crash. 
We have proposed a novel approach called the \ac{ourmethod} method for deriving \acp{ssm} that calculate the probability that a certain event, e.g., a crash, will happen in the near future given an initial situation. 
Whereas traditional \acp{ssm} are generally only applicable in certain types of scenarios, the \ac{ourmethod} method can be applied to various types of scenarios.
Furthermore, because the \ac{ourmethod} method is data-driven, the derived \acp{ssm} can be adapted to the local traffic behavior that is captured by the data.
Also, no assumptions on the driver behavior are made.
Therefore, the \ac{ourmethod} method has the potential for deriving multiple \acp{ssm} for quantifying the safety of a --- possibly automated --- vehicle.

We have illustrated that the \ac{ourmethod} method can be used to reproduce known probabilistic \acp{ssm}.
In an example, we have derived a new \ac{ssm} based on the \ac{ngsim} data set that calculates the risk of a crash in a longitudinal interaction between two vehicles.
Through several explanatory scenarios, it has been shown that the derived \ac{ssm} correctly provides a quantification of the crash risk.
We have also presented how the evaluation of the partial derivatives of the \ac{ssm} can be used to benchmark \iac{ssm} using expected risk trends.

The \acp{ssm} derived using the presented \ac{ourmethod} method can be used to warn drivers for unsafe situations and ensuring that proper attention is being paid to the road situation.
Furthermore, the derived measures can prospectively estimate the impact of newly introduced systems on traffic safety. 
A limitation of the current study is that the presented approach is only applied to longitudinal traffic interactions.
Future work involves applying the \ac{ourmethod} method for the derivation of \acp{ssm} that measure the risk of lateral traffic interactions, interactions with vulnerable road users, and interactions with multiple (different types of) traffic participants. 
Furthermore, more research is needed to investigate whether the \acp{ssm} derived by the \ac{ourmethod} method can be used to evaluate whether a driver is actively pursuing (large) safety margins.

\section*{Acknowledgement}
This research was supported in part by the SAFE-UP project (proactive SAFEty systems and tools for a constantly UPgrading road environment). SAFE-UP has received funding from the European Union’s Horizon 2020 research and innovation programme under Grant Agreement 861570.

\bibliographystyle{abbrvnat}
\bibliography{references}

\end{document}